\documentclass[10pt,twocolumn,letterpaper]{article}

\usepackage{cvpr}              

\usepackage{graphicx}
\usepackage{amsmath}
\usepackage{amssymb}
\usepackage{booktabs}

\usepackage[pagebackref,breaklinks,colorlinks]{hyperref}

\usepackage[capitalize]{cleveref}
\crefname{section}{Sec.}{Secs.}
\Crefname{section}{Section}{Sections}
\Crefname{table}{Table}{Tables}
\crefname{table}{Tab.}{Tabs.}

\def\cvprPaperID{4840} 
\def\confName{CVPR}
\def\confYear{2024}

\begin{document}

\title{Learning to Remove Wrinkled Transparent Film with Polarized Prior}

\author{Jiaqi Tang$^{1,2,3}$ \ \ \ Ruizheng Wu$^{4}$ \ \ \  Xiaogang Xu$^{5,6}$ \ \ \  Sixing Hu$^{4}$  \ \ \  Ying-Cong Chen$^{1,2,3}$\thanks{Corresponding author.}\\
$^{1}$The Hong Kong University of Science and Technology (Guangzhou) \\
$^{2}$The Hong Kong University of Science and Technology \ \ 
$^{3}$HKUST(GZ) -- SmartMore Joint Lab \\
$^{4}$SmartMore Corporation \ \ $^{5}$Zhejiang University \ \ $^{6}$Zhejiang Lab\\
{\tt\small jtang092@connect.hkust-gz.edu.cn, \{ruizheng.wu, david.hu\}@smartmore.com}\\
{\tt\small xiaogangxu@zju.edu.cn, yingcongchen@ust.hk}
}





\maketitle

\begin{abstract}
In this paper, we study a new problem, Film Removal (FR), which attempts to remove the interference of wrinkled transparent films and reconstruct the original information under films for industrial recognition systems. We first physically model the imaging of industrial materials covered by the film. Considering the specular highlight from the film can be effectively recorded by the polarized camera, we build a practical dataset with polarization information containing paired data with and without transparent film. We aim to remove interference from the film (specular highlights and other degradations) with an end-to-end framework. To locate the specular highlight, we use an angle estimation network to optimize the polarization angle with the minimized specular highlight. The image with minimized specular highlight is set as a prior for supporting the reconstruction network. Based on the prior and the polarized images, the reconstruction network can decouple all degradations from the film. Extensive experiments show that our framework achieves SOTA performance in both image reconstruction and industrial downstream tasks. Our code will be released at \url{https://github.com/jqtangust/FilmRemoval}.
\end{abstract}

\section{Introduction}
\label{sec:intro}

\begin{figure}
  \centering
  \includegraphics[width=1\linewidth]{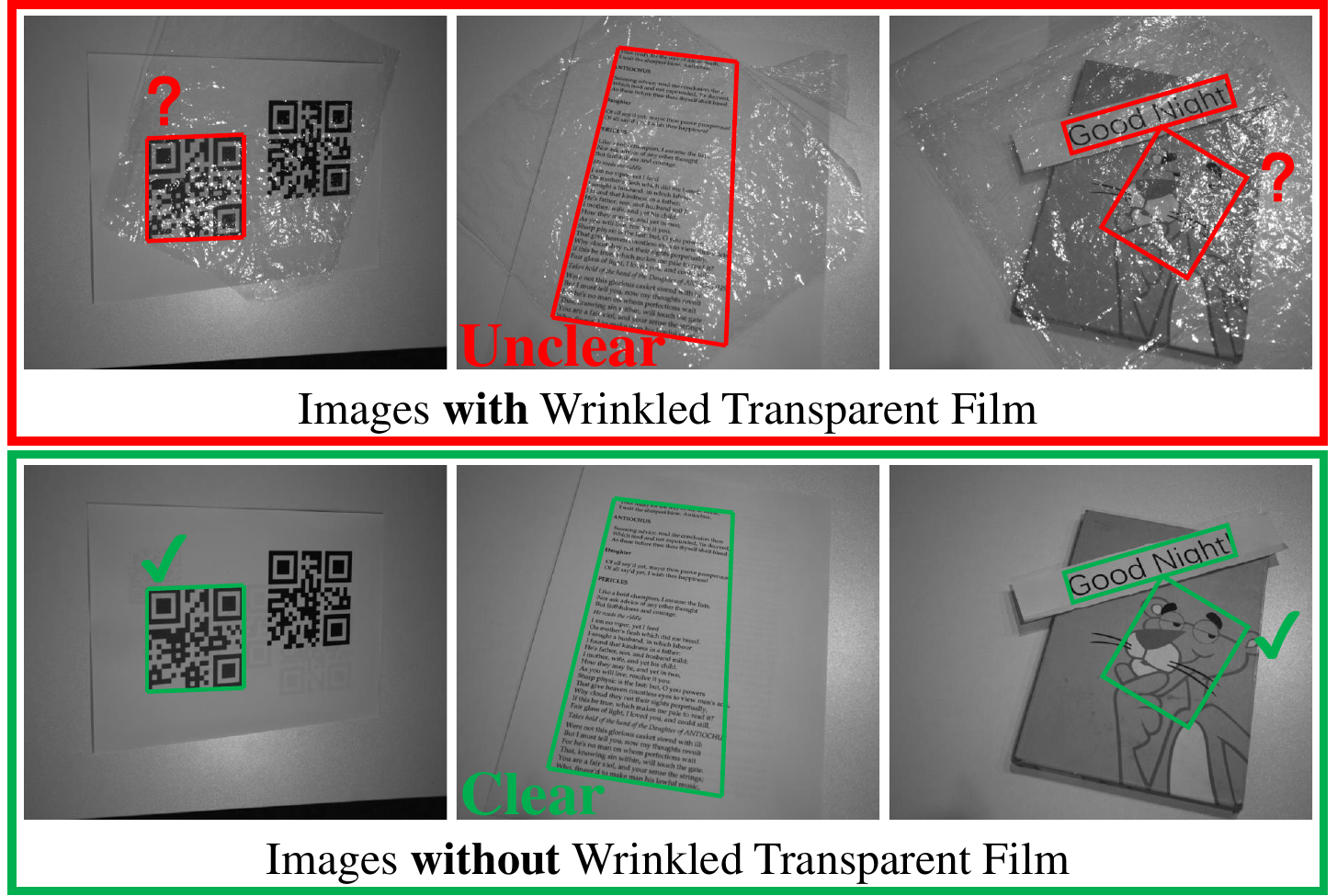}
    \vspace{-0.3in}
  \caption{
  The \textbf{\color{red}Red} box presents a challenge in industrial recognition systems, where the product information is often hidden beneath the wrinkled transparent film. The \textbf{\textcolor[rgb]{ 0,  .69,  .314}{Green}} box is the image we expect to generate, with the film layer removed. Removing the wrinkled film makes the information on industrial material clearer.}
  \label{fig:problems}
  \vspace{-0.3in}
\end{figure}

Various deep-learning-based recognition models have been employed in the industrial environment, e.g., defect detection~\cite{tabernik2020segmentation}, code recognition~\cite{shen2021blind}, etc. However, the model failures would sometimes happen due to the insufficient robustness~\cite{yin2019fourier} towards different perturbations~\cite{huang2019enhancing,sayed2021improved,kim2018diminishing,yang2020raindrop,wang2008face}. The wrinkled transparent film is one of them, which is usually covered or packaged on industrial materials or products for protection. Its interference could cause the failure of varying downstream tasks, e.g., text OCR and QR code recognition, in Fig.~\ref{fig:problems}. Regarding the wide usage of such films in industrial scenarios, it is worth opening the research direction to remove these films from images.

\begin{figure*}
  \centering
  \includegraphics[width=1.0\linewidth]{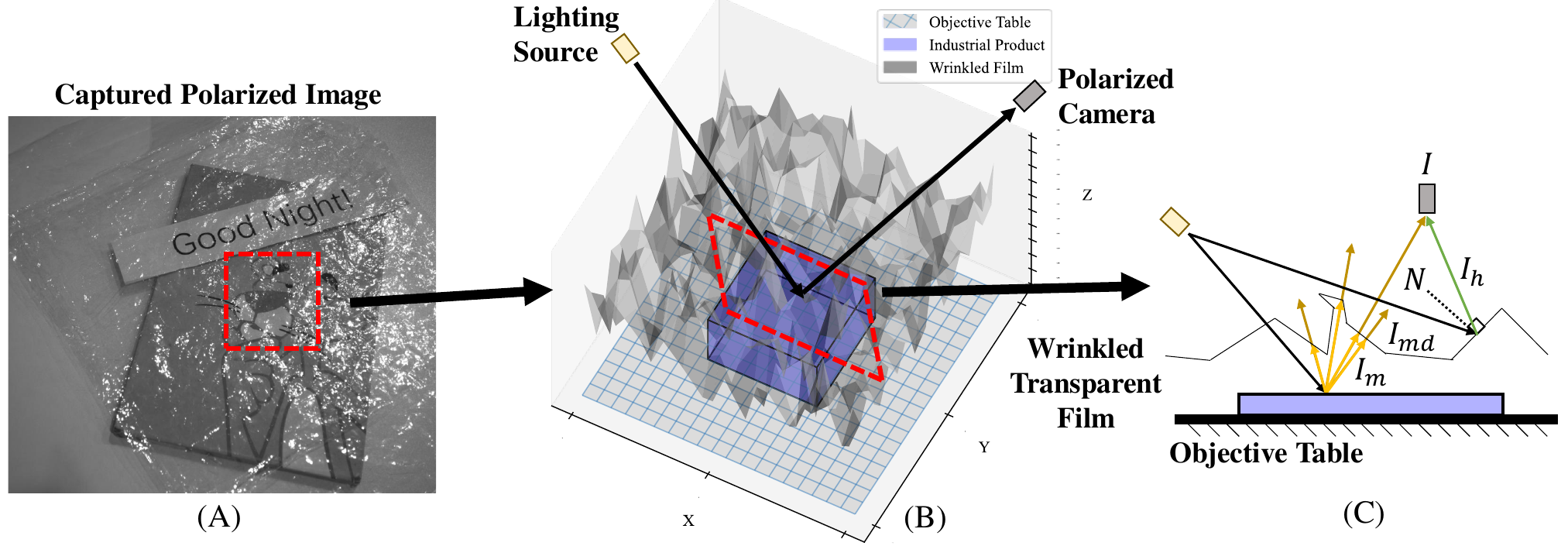}
  \vspace{-0.3in}
  \caption{Wrinkled Transparent Film Model. (A) The polarized image. (B) The 3D physics model of the local region. The light is reflected through the transparent wrinkled film and captured by the polarization camera. (C) The light path diagram. Polarized cameras capture two components: \textbf{\textcolor[rgb]{ .439,  .678,  .278}{Specular Reflection}} ($I_h$), and \textbf{\textcolor[rgb]{  .746, 0.565, 0.0}{Diffuse Reflection}} ($I_{md}$). The \textbf{\textcolor[rgb]{  1, .753, 0}{Original Diffuse Reflection}} ($I_{m}$) would be interfered by various degradations ($I_{md} - I_{m}$) from the film.}
  \label{fig:fm}
  \vspace{-0.25in}
\end{figure*}

For the first time, we address a novel problem of wrinkled transparent Film Removal (FR), which aims to remove the transparent film and reveal the hidden information, benefiting the robustness of the industrial downstream models. 

Although some solutions have attempted to remove surface highlight~\cite{wen2021polarization, Wang2012RemovalOT, fu-2021-multi-task}, 
they have not accurately modeled the imaging of the wrinkled transparent film.
Except for the highlight, they cannot remove the effects of other various degradations from the film, e.g., light transmittance and material texture. Therefore, they cannot remove the transparent film thoroughly.

In this paper, we explicitly model the imaging of wrinkled transparent film into two parts: the specular highlight $I_{h}$ from the film,
and the diffuse reflection $I_{md}$ from the materials under the film, as shown in Fig.~\ref{fig:fm}. The diffuse reflection can be influenced by the properties of the film. Thus, it is divided into the original component ($I_m$) and various degradations ($I_{md}-I_m$). Therefore, our objective is to decouple the specular highlight and other degradations caused by the film and reconstruct the original diffuse reflection.

We build an end-to-end framework for decoupling two different degradations in the Wrinkled Transparent Film Model (Fig.~\ref{fig:fm}), which consists of the prior estimation and reconstruction network.
We notice that specular highlight is significantly related to polarization angles, while other parts are not. Based on this observation, we use an Angle Estimation Network to learn the polar angle corresponding to the minimized specular highlight, which is driven by a Polarization-based Location Model.
Images with minimized specular highlights are set as the priors for the later reconstruction network to remove all degradations.

There is currently no suitable dataset for the FR problem, since most of the existing datasets~\cite{wen2021polarization,fu-2020-learn-detec,meka2018lime} are targeted towards specular reflection removal only.
Also, most datasets do not consider the real industrial environment. Therefore, we build a new dataset that consists of paired images covered by the film and the uncovered ground truth in the industrial optical photography system. Moreover, it's proved that the specular highlight from the film can be effectively captured by the polarization camera~\cite{wolff1989using,wen2021polarization,nayar1997separation}, which is now cheap to be installed in industrial systems~\cite{Rebhan2019PrincipleIO}. 
Thus, we follow the collection pipeline of existing polarization datasets~\cite{de1994brewster,kanseri2023degree}, and capture each object with four polarized images under four polar directions in one shot. It's empirically proved by our experiments that with the input of these polarization clues, networks can better locate the specular highlight and recover the hidden information under the film.

Extensive experiments prove that our designed network achieves SOTA performance in the FR problem.
Our contributions are summarized as follows:
\begin{itemize}
\vspace{-0.1in}
\item To the best of our knowledge, we are the first to address the new problem of Film Removal (FR), which aims to remove the whole wrinkled transparent film in industrial scenarios.
\vspace{-0.1in}
\item To solve FR, we model the wrinkled film physically and propose an end-to-end reconstruction network for FR with a learnable polarization-based prior, which helps the network locate the specular highlight reflection in the film.
\vspace{-0.1in}
\item We also build a new practical dataset in the real industrial optical photography system, which contains various polarized image pairs with and without the film.
\vspace{-0.3in}
\item Extensive experiments are conducted to prove the effectiveness of our dataset and method, which achieves SOTA performance in both image reconstruction and downstream industrial tasks.
\end{itemize}
\vspace{-0.2in}
\section{Related Work}
\label{sec:related}
\subsection{Polarization Model and Application}
\label{sec:re-polar}
Polarization refers to the property of the transverse wave oscillating in different directions. Since light is a kind of wave, this phenomenon describes the distribution of light waves in all directions~\cite{atkinson2020polarized}. Conventional cameras or human eyes are insensitive to polarization. Thus, polarization is often used as a way to supplement additional visual information. It is often used in a wide range of fields such as optics~\cite{ulrich1979polarization}, materials science~\cite{sundar1992electrostriction}, and physics~\cite{wolfenstein1956polarization,gold1952polarization}, etc.

In the field of computer vision, polarization provides different angles of view hence allowing more efficient interpretation of complex scenes. In recent years, polarization has been widely used for complex tasks such as Integral Imaging~\cite{xiao2012three}, Rendering~\cite{baek2020image}, 3D Shape~\cite{zhu2019depth}, Segmentation~\cite{liang2022multimodal}, and Reflection Removal~\cite{lyu2019reflection,wen2021polarization,lei2020polarized}, etc.

In the real world, since natural light is mixed with multiple wavelengths, its refraction time is different when it is injected into an optically active material, thus a phase shift occurs. A physical description of this phenomenon is elliptically polarized light (\ie, partially polarized light), as shown in Eq.~\eqref{r1}.
\vspace{-0.1in}
\begin{equation}
\vspace{-0.1in}
    \left\{
    \begin{array}{ll}
    E_x=E_{x0}\cos{(\omega t)} \\
    E_y=E_{y0}\cos{(\omega t -\sigma)}
    \end{array}
  \right. ,
    \label{r1}
\end{equation}
where ${({x}},{{y})}$ is the Cartesian basis in the space of Jones vectors, $E_{x}$ and $E_{y}$ is the component of the light on the basis, ${({x}},{{y})}$. $\omega$ is the frequency and $\sigma$ is the phase difference of ${{E_x}}$ and ${{E_y}}$. $E_{x0}$ and $E_{y0}$ are the field strength of a pair of orthogonal waves. Eq.~\eqref{r1} describes the vibration of the polarized light in different angles, $t$. Based on this model, our solution only considers a simplified situation where
$\sigma=90^{\circ}$. Eq.~\eqref{r1} can be simplified to Eq.~\eqref{r3},
\vspace{-0.15in}
\begin{equation}
\vspace{-0.1in}
    \frac{E_{x}^{2}}{E_{x0}^{2}}+\frac{E_{y}^{2}}{E_{y0}^{2}}=1.
    \label{r3}
    \vspace{-0.05in}
\end{equation}

\begin{figure}
  \centering
  \includegraphics[width=0.5\linewidth]{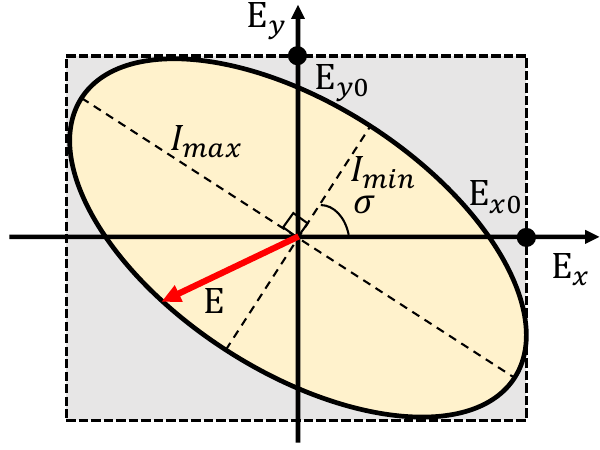}
  \vspace{-0.2in}
  \caption{Model of elliptically polarized light. $E$ represents polarized light at any angle, which can be calculated by this model. $I_{max}$ and $I_{min}$ are two components indicating the maximum and minimum intensity of elliptically polarized light.}
  \vspace{-0.26in}
  \label{fig:phy}
\end{figure}


\subsection{Specular Highlight Removal via Polarization}
Since the film includes unpredictable specular reflection, our task includes the procedure to remove these degradations. There have been several solutions to remove specular reflection by polarization information. Nayar et al.~\cite{nayar1997separation} first used polarization to determine the color of the specular component to separate the interfaces. Then, Umeyama et al.~\cite{umeyama2004separation} adopted independent component analysis to separate the diffuse and specular reflection components of surface reflection. Zhang et al.~\cite{zhang2011reflection} considered the effect of polarization angle and attempted to get the appropriate global angle using Newton's method, but this method did not make full use of the local information. 
Wen et al.~\cite{wen2021polarization} separated specular reflection regions by using image chromaticity. 

Although conventional methods are available for removing specular highlight reflections, they are not able to accurately model the imaging of wrinkled transparent films, and thus do not adequately address the problem of eliminating all degradations from wrinkled transparent films.



\vspace{-0.03in}
\section{Dataset}
\label{sec:data}
\vspace{-0.03in}

While some datasets exist for the removal of specular reflection~\cite{Li_2017_CVPR,wen2021polarization,lei2022categorized}, there is currently no dataset for film removal in the industrial environment.
As shown in Fig.~\ref{fig:fm}, the characteristics of the specular highlight information from the film can be effectively modeled by the polarized image.
Leveraging this polarization information can significantly enhance image reconstruction.
Therefore, we construct a paired dataset based on polarized images. 
Polarized images can capture light amplitude information from wrinkled films at different angles, and making full use of this information will significantly facilitate the decoupling of degradation information occurring from the film.
\vspace{-0.02in}
\subsection{Industrial Optical Photography Pipeline}
\begin{figure}
  \centering
  \includegraphics[width=1\linewidth]{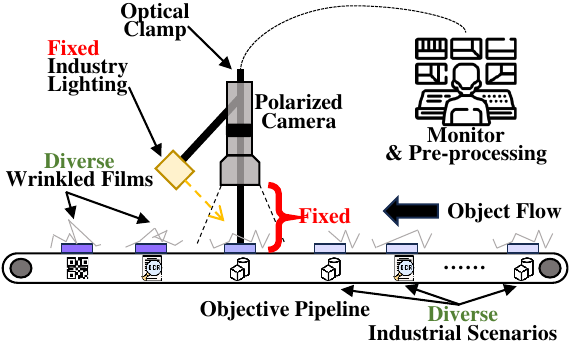}
  \vspace{-0.33in}  
  \caption{Prototype of industrial optical photography pipeline. We have built an optical pipeline for capturing the dataset in the industrial environment. 
  As objects traverse the objective pipeline, the polarizing camera captures images continuously.
  Subsequently, the acquired data is sent to the monitor for pre-processing.}
  \label{fig:system}
  \vspace{-0.26in}
\end{figure}

\vspace{-0.01in}
\label{ps}
Fig.~\ref{fig:system} illustrates the prototype of our pipeline. 
Within this industrial pipeline, we maintain a consistent posture and angle for both the camera and lighting resources. The objective pipeline sequentially passes different detected objects under the camera. 

To effectively capture optical information in multiple polarization directions within a single image, we employ the HIKROBOT MV-CH050-10UP camera\footnote{\url{https://www.hikrobotics.com/cn/machinevision/productdetail?id=3886}}, which integrates the Sony IMX250MZR CMOS sensor\footnote{\url{https://www.sony-semicon.com/en/products/is/industry/polarization.html}}. This sensor contains four different angles of polarization (${0}^{\circ}$, ${45}^{\circ}$, ${90}^{\circ}$, and ${135}^{\circ}$) as Bayer pattern, allowing us to capture four polarization angles in high definition with a single shot.

Finally, there is a monitor that controls the camera shutter and adjusts camera parameters. We utilize auto-exposure and auto-focus strategies to set the appropriate focal length, exposure time, and ISO for each specific scenario automatically. This is essential since the thickness and surface of the products on the industrial line may vary, requiring slight adjustments in camera parameters to ensure image quality.
\vspace{-0.02in}
\subsection{Capturing Polarized Images}
\vspace{-0.01in}
When capturing polarized images, the imaging system initially captures an image of the ground truth uncovered by the film, $I^{raw}_{gt}$. Subsequently, with the complex transparent film placed over the ground truth, we capture the image $I^{raw}_{input}$ that needs to be recovered.

\vspace{-0.02in}
\subsection{Data Diversity and Robustness}
\vspace{-0.01in}
To build a diverse and robust dataset, we follow the rules in industrial manufacturing. As depicted in Fig.~\ref{fig:system}, the current industrial pipeline includes 315 dynamic industrial scenarios, which can be categorized into three types: QR codes, text, and products. To enhance the diversity, we have different films with diverse material properties, coverage areas, film thicknesses, and levels of wrinkling. The film exhibits significant variability across each scenario.

On the other hand, to ensure the stability of the industrial imaging pipeline, we maintained a consistent intensity level for the industrial light source and fixed the distance between the camera and the object flow. This helps to minimize the influence of errors external to the industrial system.


\vspace{-0.02in}
\subsection{Prepossessing}
\vspace{-0.01in}
Each pixel of the image captured by the polarization sensor is represented by four-pixel dots, each corresponding to the intensity of light at four distinct polarization angles.
We first need to decompose it into 4 subgraphs with different angles, and then we restore it to its original resolution using edge-aware residual interpolation (EARI) demosaicking~\cite{morimatsu2020monochrome}. This process is described in Eq.~\eqref{1:cp}.
\vspace{-0.12in}
\begin{equation}
\vspace{-0.12in}
\begin{aligned}
    \{ I_{gt}^{0},I_{gt}^{45},I_{gt}^{90},I_{gt}^{135} \}&=M ( F_{d}(I^{raw}_{gt})), \\
    \{ I_{input}^{0},I_{input}^{45},I_{input}^{90},I_{input}^{135} \}&=M ( F_{d}(I^{raw}_{input})),
\end{aligned}
\label{1:cp}
\end{equation}
where $F_d(\cdot)$ is the decomposing operator, $M(\cdot)$ is EARI demosaicking. Taking $I^{raw}_{gt}$ and $I^{raw}_{input}$ as inputs, we obtain full-resolution polarized images with four different angles. 

To generate the ground truth image, we follow the standard procedure from the polarized image processing library, polanalyser\footnote{\url{https://github.com/elerac/polanalyser/wiki}}. Firstly, we introduce stokes parameters~\cite{bickel1985stokes}. According to this physical model, the first stoke parameter $S_0$ describes the total intensity of the optical beam, and it can be calculated by Eq.~\eqref{s0}.
\vspace{-0.12in}
\begin{equation}
\vspace{-0.08in}
    \label{s0}
    S_0=E_{x0}^{2}+E_{y0}^{2}=I^{x}+I^{y},
\vspace{-0.02in}
\end{equation}
where $I^{x} \perp I^{y}$. $E_{x0}^{2}$ and $E_{y0}^{2}$ are the field strength of a pair of orthogonal waves in Fig.~\ref{fig:phy}, which can be calculated by a pair of orthogonal polarization components, $I^{x}$ and $I^{y}$.
Subsequently, we can calculate one ground truth $I_{gt}$ as 
\vspace{-0.12in}
\begin{equation}
\vspace{-0.12in}
    \label{cal}
    I_{gt}=G(\frac{S0}{2})={(\frac{I_{gt}^{0}+I_{gt}^{45}+I_{gt}^{90}+I_{gt}^{135}}{4})}^{\frac{1}{\gamma}},
\vspace{-0.03in}
\end{equation}
where $G(\cdot)$ is a gamma correction function, and we empirically set the gamma value, $\gamma$, to $2.2$.

\vspace{-0.035in}
\subsection{Training and Testing}
\vspace{-0.035in}
During training, our network mixed all the scenes for training, and the final network is applicable to data from all scenarios. Besides, to ensure the robustness and generalization of our dataset, we adopt 10-fold cross-validation~\cite{refaeilzadeh2009cross} to evaluate the results. The dataset is divided into ten parts, and nine of them are used as training data and one as test data in turn. Each test will yield a corresponding accuracy rate, which is then averaged as the final accuracy.

\vspace{-0.05in}
\section{Method}
\vspace{-0.01in}
\begin{figure}
  \centering
  \small
  	\newcommand\widthface{0.5}
	\begin{tabular}{ccc}
	    Specular Highlight, $I_{h}$& \textbf{\textcolor[rgb]{ 1,  1,  1}{}} &  \textbf{\textcolor[rgb]{ 1,  1,  1}{...}} Other Degradations, $I_{d}$
	\end{tabular}
        \resizebox{1.0\linewidth}{!}{
        \includegraphics[width=\widthface\textwidth]{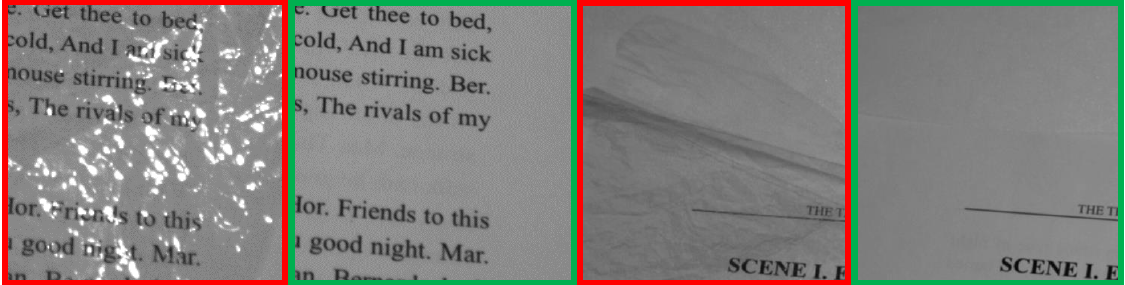}}
 \vspace{-0.3in}
  \caption{Two Decoupling Components. Specular highlight, $I_{h}$ and other degradations, $I_{d}$. The \textbf{\textcolor[rgb]{ 1,  0,  0}{Red}} box shows the degradations, the \textbf{\textcolor[rgb]{ 0,  .69,  .314}{Green}} box is the Ground Truth.}
  \vspace{-0.27in}
  \label{fig:dg}
\end{figure}
\label{sec:method}
\vspace{-0.05in}
Although some conventional methods have used polarization information to remove surface specular highlight reflection, they assume the light intensity as a \textit{binary} composition~\cite{wen2021polarization}, i.e, transmission and reflection. However, the light intensity from the wrinkled film is more complex, which is not only influenced by film surface highlight but also mixed with various degradations, e.g., light transmittance and material texture.

\begin{figure*}
  \centering
  \includegraphics[width=1.0\linewidth]{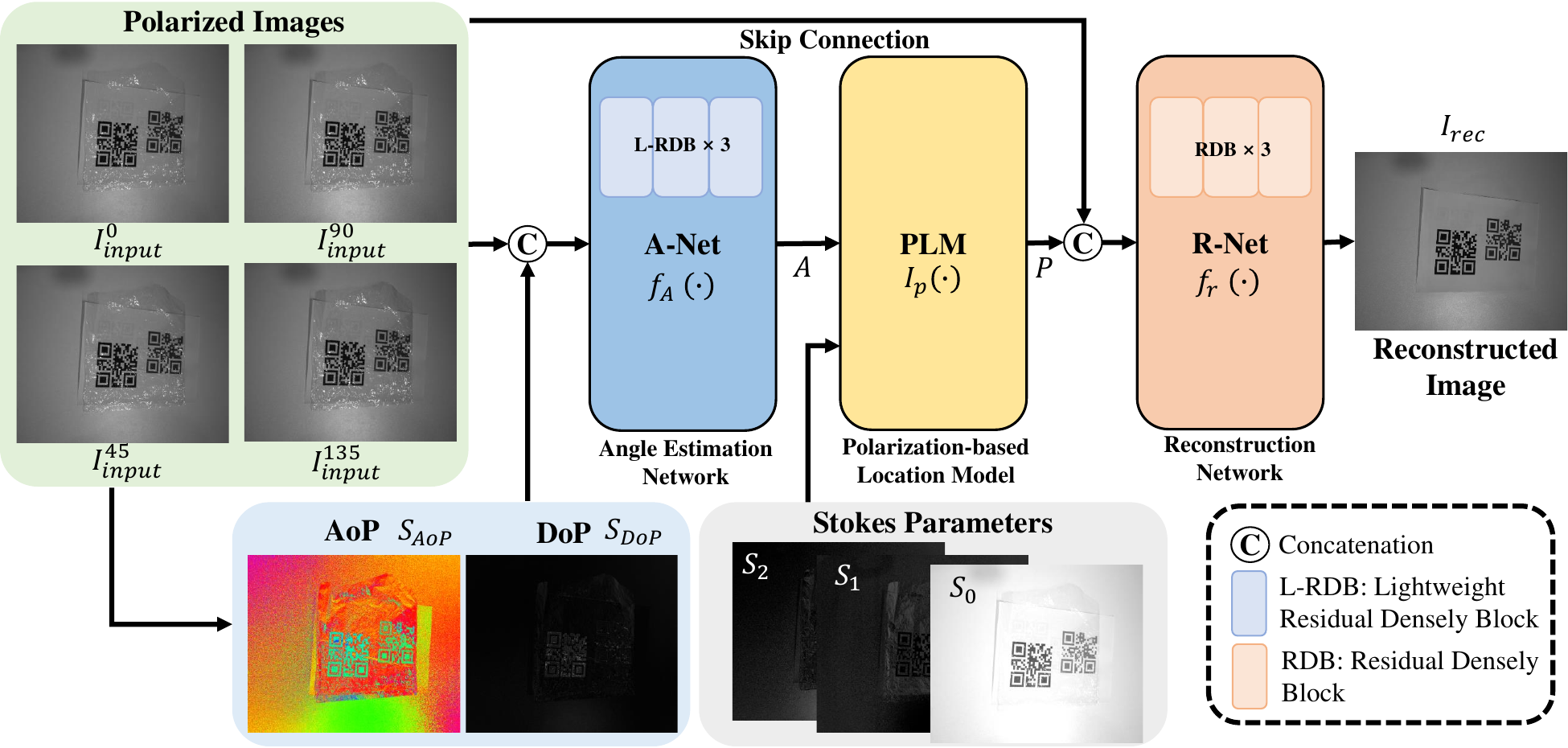}
  \vspace{-0.28in}
  \caption{Overall framework. The polarized images, AoP and DoP are fed into Angle Estimation Net (A-Net), denoted as $f_{A}$, for estimating the angle, $A$. Subsequently, the Polarization-based Location Model (PLM), represented as $I_p$, takes $A$ as input to estimate the image prior, $P$. This prior provides important highlight location information for the reconstruction network. Finally, the reconstruction network (R-Net) is trained to restore the original diffuse reflection in industrial materials.}
\vspace{-0.25in}
  \label{fig:method}
\end{figure*}

In this section, we first model the wrinkled transparent film physically in Fig.~\ref{fig:fm}, which is our method's motivation. Then, we use an end-to-end network for reconstructing the original information in Fig.~\ref{fig:method}. Besides, since the highlight region is more tricky to recover, we build a polarization-based prior into the end-to-end 
framework to assist in locating highlight regions.

\vspace{-0.035in}
\subsection{Modelling the Wrinkled Transparent Film}
\vspace{-0.035in}
\label{sec:reflection}

Based on the Industrial Optical Photography System in Sec.~\ref{ps}, the light intensity for materials covered with the wrinkled transparent film will be captured by the polarized camera in Fig.~\ref{fig:fm}(A)(B). The captured light intensity consists of two parts, i.e., the specular highlight reflection of the film's polarized regions~\cite{monzon2017anisotropy, sinichkin2010effect}, the diffuse reflection from the materials under the film, which could be influenced by light transmittance, the texture of the film in Fig.~\ref{fig:fm}(C). Such a composition is written in Eq.~\eqref{model}.
\vspace{-0.14in}
\begin{equation}
\vspace{-0.14in}
    \begin{split}
    I &= I_{md} + I_{h} = I_{m} + I_{d} + I_{h},
    \label{model}
    \end{split}
\end{equation}
where $+$ denotes the linear superposition of different light components, $I$ is the light intensity captured by the camera, $I_{md}$ is the diffuse reflection component of the material through various film degradations, $I_h$ is the specular reflection part from the film's highlighted regions. Then, $I_{md}$ can be decoupled to two parts, $I_{m}$ and $I_{d}$. $I_{m}$ is the original diffuse reflection component of the material and $I_{d}$ is caused by other various degradations through film. In Fig.~\ref{fig:fm}(C), the optical path diagram illustrates this process.

Based on Eq.~\eqref{model}, the FR task can be implemented by retaining the information of $I_{m}$ and decoupling $I_{h}$ and $I_{d}$ (both of them are caused by the film layer). This is expressed in Eq.~\eqref{optim}, as
\vspace{-0.16in}
\begin{equation}
\vspace{-0.12in}
    I_{m}=I- I_{h}- I_{d},
    \label{optim}
\end{equation}
where $-$ denotes the decoupling operator. Fig.~\ref{fig:dg} visualizes these two components for decoupling in our FR task.

Based on this model, our whole framework is based on an end-to-end reconstruction network for decoupling these two parts in Sec.~\ref{sec:recon}. Before that, to support the network for decoupling $I_{h}$, we estimate a polarized prior for locating the highlight regions in Sec.~\ref{sec:estimate}.

\vspace{-0.035in}
\subsection{Estimating a Polarized Prior for Locating $I_h$}
\vspace{-0.035in}
\label{sec:estimate}

To decouple the specular reflection components, it is better to locate the highlight regions on the surface of the wrinkled film, which provides a prior to facilitate the decoupling network. However, these regions are hard to predict since images are captured in variable scenarios. To this end, we introduce the polarization information for this problem. 

Based on Fresnel’s theory~\cite{Winthrop1965TheoryOF}, the specular reflection component of optically active materials is an elliptically polarized light, which changes under different angles of polarization orientation, while the rest of the components remain almost constant. It can be employed in estimating $I_{h}$, which is a specular reflection component. Thus, we propose a polarized prior $P$, which is represented as the optimized $I$ with the minimized $I_{h}$, as shown in Eq.~\eqref{P}. 
The difference between $P$ and the input $I$ indicates the regions of $I_{h}$.
\begin{figure}
  \centering
	\Huge
        \vspace{-0.05in}
	\newcommand\widthface{0.45}
	\resizebox{1.0\linewidth}{!}{
	\begin{tabular}{ccc}
	    Prior & Location (Highlighted) & Location\\
	    \includegraphics[width=\widthface\textwidth]{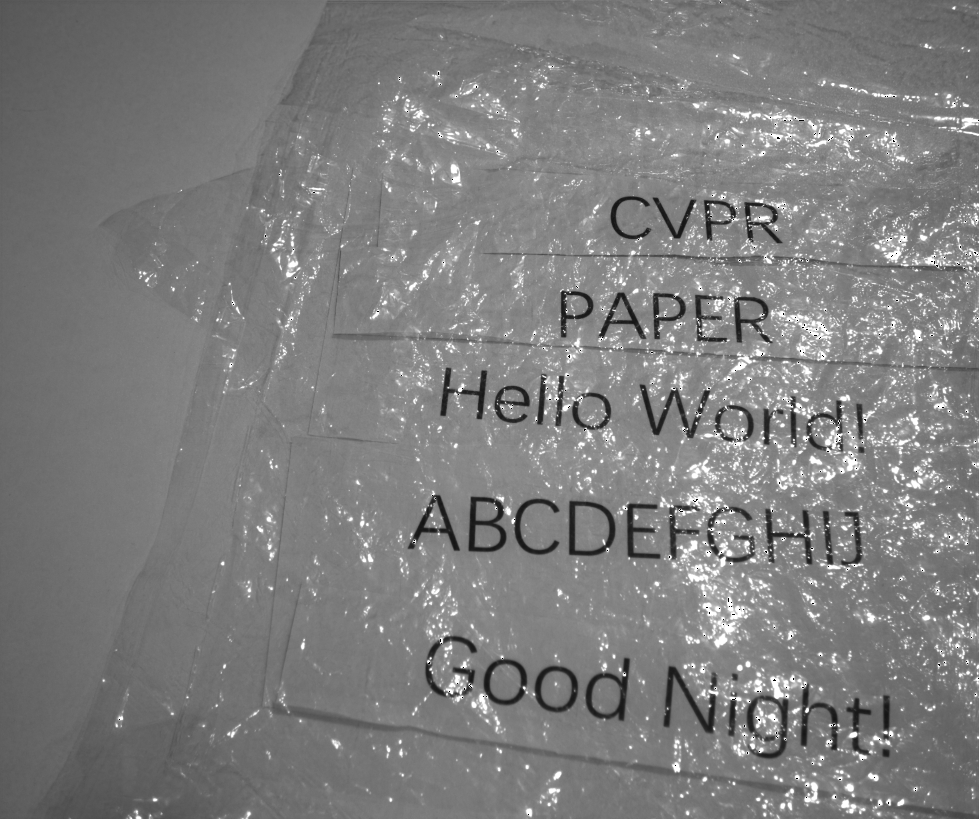} &
	    \includegraphics[width=\widthface\textwidth]{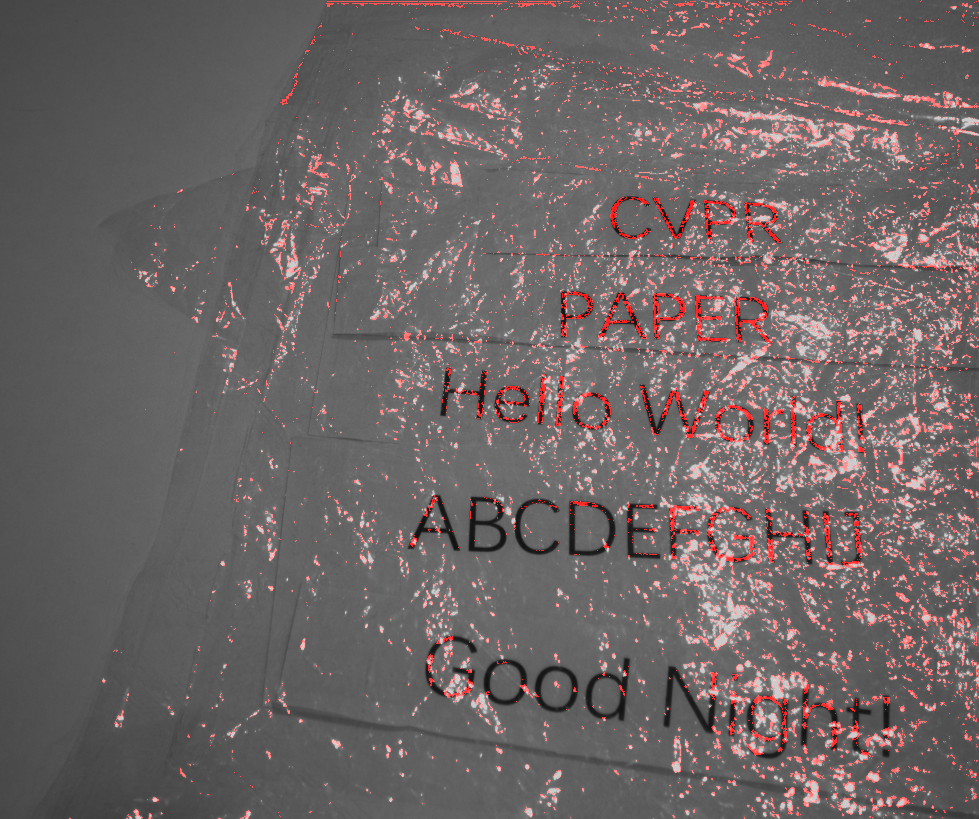} &
		\includegraphics[width=\widthface\textwidth]{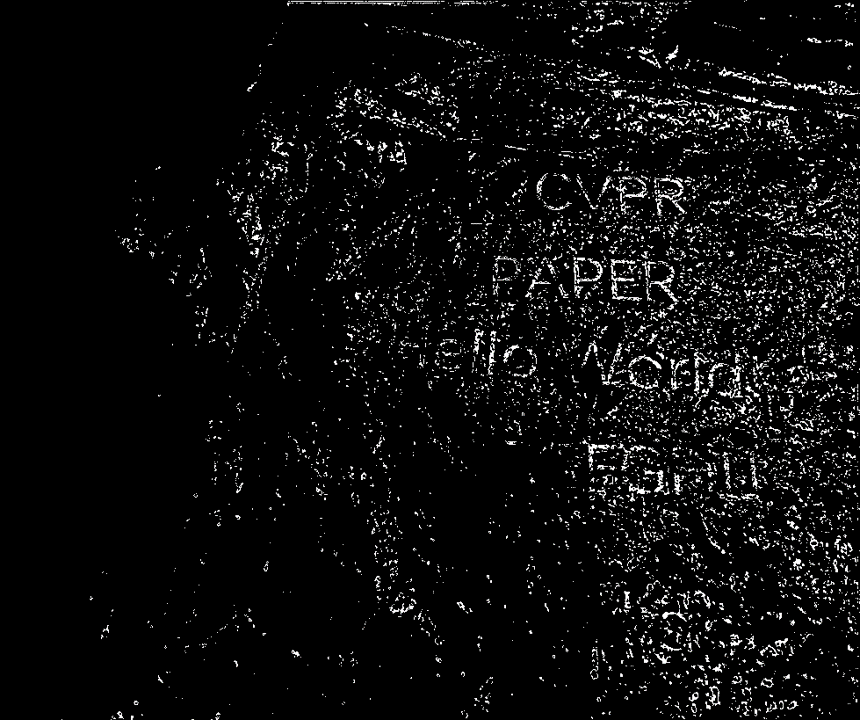}
	\end{tabular}}
 \vspace{-0.18in}
  \caption{Location of specular highlight. The $P$ is the polarized prior. We can calculate the location of highlight by $I_h- \min I_h$.}
   \vspace{-0.23in}
  \label{fig:P}
\end{figure}
\vspace{-0.1in}  
\begin{equation}
\vspace{-0.1in}  
    P = I_{m}+I_{d}+\min I_{h}.
    \label{P}
\end{equation}

Eq.~\eqref{P}, \ie, the polarized version of Eq.~\eqref{model}, can be acquired with Malus’s Law~\cite{malus1810theorie} and the elliptical polarization model in Sec.~\ref{sec:re-polar}. We use a pair of orthogonal maximum and minimum components and the angle variable $\theta \in [0,2\pi)$ to rewrite Eq.~\eqref{P} as
\vspace{-0.1in}
\begin{equation}
\vspace{-0.1in}
    I_{h} = I_p(\theta) =I_{max}\cos^2{\theta}+I_{min}\sin^2{\theta},
    \label{xymalus}
\end{equation}
where $I_p(\theta)$ is the Polarization-based Location Model to describe the polarized light in the angle of $\theta$, $I_{max}$ and $I_{min}$ is a pair of orthogonal maximum and minimum components in Fig.~\ref{fig:phy}. Given the input data $\{I_{input}^{0},I_{input}^{45},I_{input}^{90},I_{input}^{135} \}$, $I_{max}$ and $I_{min}$ can be computed by the Stokes parameters~\cite{bickel1985stokes}, from Eq.~\eqref{maxmin} and Eq.~\eqref{S}.
\vspace{-0.11in}
\begin{equation}
\small
\vspace{-0.1in}
    \begin{split}
            I_{max}=S_0 + \frac{\sqrt{S_1^2+S_2^2}}{2},
        I_{min}=S_0 - \frac{\sqrt{S_1^2+S_2^2}}{2},
    \end{split}
    \label{maxmin}
\end{equation}
\begin{equation}
\vspace{-0.05in}
\small
\begin{split}
        S_0=E_{x0}^{2}+E_{y0}^{2}=I^0_{input}+I^{90}_{input},\\
    S_1=E_{x0}^{2}-E_{y0}^{2}=I^0_{input}-I^{90}_{input},\\
    S_2=E_{a0}^{2}-E_{b0}^{2}=I^{45}_{input}-I^{135}_{input},
\end{split}
    \label{S}
\end{equation}
where ${({a}},{{b})}$ is the Cartesian basis rotated by 45° in the space of ${({x}},{{y})}$, and $E_{a0}^{2}$ as well as $E_{b0}^{2}$ are the field strengths under this basis.

Since $I_{h}$ is the only polarized component that is determined by $\theta$, $P$ in Eq.~\eqref{P} can also be formulated as Eq.~\eqref{xymalusmin}.
\vspace{-0.14in}
\begin{equation}
    \begin{split}
    P &= I_{m}+I_{d}+\min I_{h} \\&= I_{m}+I_{d}+\min_{\theta} I_p(\theta)\\
               &= I_{m}+I_{d}+\min_{\theta} (I_{max}\cos^2{\theta}+ I_{min}\sin^2{\theta}).
    \label{xymalusmin}
    \end{split}
\end{equation}
\begin{table*}[t]
  \centering
  \Large
    \resizebox{1.0\linewidth}{!}{
    \begin{tabular}{l|c|cccccccccc|cc}
    \toprule
    \multicolumn{1}{c}{}  &       & K1    & K2    & K3    & K4    & K5    & K6    & K7    & K8    & K9    & K10   & $\mu \uparrow$ & $\sigma \downarrow$ \\
    \midrule
    SHIQ~\cite{fu-2021-multi-task}& PSNR & 23.47 & 22.11 & 21.95 & 21.69 & 21.77 & 21.03 & 20.86 & 20.46 & 21.10 & 21.31 & 21.58 & 0.64 \\ 
    &SSIM & 0.7899 & 0.7640 & 0.7416 & 0.7439 & 0.7459 & 0.7465 & 0.7499 & 0.7412 & 0.7465 & 0.7300 & 0.7499 &  $2.41\times 10^{-4}$ \\ 
    
    \midrule
     Polar-HR~\cite{wen2021polarization}& PSNR & 23.31 & 22.80 & 22.13 & 21.58 & 21.94 & 22.00 & 22.03 & 21.99 & 22.18 & 21.95 & 22.19 & 0.22 \\ 
& SSIM & 0.7642 & 0.7421 & 0.7220 & 0.7099 & 0.7064 & 0.7098 & 0.7128 & 0.7017 & 0.7102 & 0.6968 & 0.7176 & $3.80\times 10^{-4}$ \\ 
          
    \midrule
    Uformer~\cite{wang2022uformer} & PSNR  & 31.85 & 31.95 & 31.39 & 31.19 & 31.81 & 32.04 & 31.68 & 31.98 & 31.85 & 31.01 & 31.68 & \textbf{0.11} \\
          & SSIM  & 0.9519 & 0.9456 & 0.9371 & 0.9364 & 0.9434 & 0.9421 & 0.9438 & 0.9435 & 0.9457 & 0.9363 & 0.9426 & $2.17\times 10^{-5}$ \\
    \midrule
    Restormer~\cite{Zamir2021Restormer} & PSNR  & 34.35 & 35.02 & 34.44 & 33.71 & 34.88 & 35.13 & 34.31 & 34.33 & 34.51 & 32.49 & 34.32 & 0.52 \\
          & SSIM  & 0.9771 & 0.9770 & 0.9721 & 0.9678 & 0.9757 & 0.9746 & 0.9742 & 0.9741 & 0.9759 & 0.9633 &  0.9731 & $1.75\times 10^{-5}$ \\
    \midrule
    Ours & PSNR  & \textbf{36.76} &\textbf{ 37.29} & \textbf{36.62} & \textbf{35.12} & \textbf{36.93} & \textbf{37.21} & \textbf{36.24} & \textbf{36.67} & \textbf{36.94} & \textbf{35.02} & \textbf{36.48} & 0.57\\ 
         &  SSIM & \textbf{0.9852} & \textbf{0.985}\textbf{9} & \textbf{0.9822} & \textbf{0.9767} & \textbf{0.9845} & \textbf{0.9833} & \textbf{0.9836} & \textbf{0.9830} & \textbf{0.9850 }& \textbf{0.9749} & \textbf{0.9824} & $\textbf{1.23}\times\textbf{10}^{-\textbf{5}}$\\
    \bottomrule
    \end{tabular}%
    }
\vspace{-0.13in}
\caption{Quantitative evaluation in 10-fold cross-validation. The K-$I$ indicates the $I_{th}$ fold.}
\vspace{-0.16in}
\label{cross}
\end{table*}%
Different pixels in one image correspond to varying optimized values for $\theta$. Thus, we estimate pixel-wise $\theta$ with a learning-based network, obtaining the angle map $A$.
The input includes images with polarization information, i.e., $I_0$, $I_{45}$, $I_{90}$, and $I_{135}$.
In addition, our input also includes two essential physics statistics, angle of polarization (AoP) and degree of polarization (DoP)~\cite{mcmaster1961matrix}. The AoP provides information about the object's surface normal, which helps to analyze the difference in the surface structure between the object and film. The DoP offers information on the intensity of polarized light, so it can facilitate the network to utilize the polarized light accurately. Both of them can promote the model to learn the appropriate angle better. Eq.~\eqref{angle} describes this procedure.
\vspace{-0.12in}
\begin{equation}
\vspace{-0.12in}
\resizebox{0.85\linewidth}{!}{
$A=f_A (I_{input}^{0} \oplus
I_{input}^{45}\oplus
I_{input}^{90}\oplus
I_{input}^{135}\oplus
S_{AoP}\oplus
S_{DoP})$},
    \label{angle}
\end{equation}
where, $A \in \mathbb{R}^{h\times w \times 1}$ is the pixel-wise angle map, $S_{AoP}\in \mathbb{R}^{h\times w \times 1}$ and $S_{DoP}\in \mathbb{R}^{h\times w \times 1}$ indicate the AoP and DoP map respectively,
$\oplus$ is the concatenation operator, and $f_A(\cdot)$ is angle estimation network. 
The network structure employs a lightweight Residual Dense Network~\cite{zhang2020residual} with a large perception field to capture more global information. 

After obtaining the angle map $A$, we can get the prior with minimized $I_{h}$ by Eq.~\eqref{vpf}.
\vspace{-0.12in}
\begin{equation}
\vspace{-0.12in}
    P = I_{m}+I_{d}+I_{max}\cos^2{A}+I_{min}\sin^2{A},
    \label{vpf}
\end{equation}
where, $P \in \mathbb{R}^{h\times w \times 1}$. Fig.~\ref{fig:P} visualize the Prior, $P$, and the location of $I_h$ (highlighted). We can accurately estimate the location of the specular highlight.
\vspace{-0.03in}
\subsection{Reconstructing $I_{m}$ with a Prior}
\vspace{-0.03in}
\label{sec:recon}
In this section, we set a reconstruction network $f_r$ to decouple both $I_{d}$ and $I_{h}$, with the input of $I$ (i.e., $I_{input}^{0}$,$I_{input}^{45}$,$I_{input}^{90}$, and $I_{input}^{135}$) and $P$.
The obtained prior $P$ from Sec.~\ref{sec:estimate} can already provide the estimation of $I_{h}$ by comparing with $I$. Thus, $f_r$ can be focused on the assessment of $I_{d}$.
The reconstruction network is implemented as a common Residual Dense Network~\cite{zhang2020residual}.
The reconstruction process can be expressed as Eq.~\eqref{recon}.
\vspace{-0.12in}
\begin{equation}
\vspace{-0.12in}
    I_{rec}=f_r(I_{input}^{0}\oplus
I_{input}^{45}\oplus
I_{input}^{90}\oplus
I_{input}^{135}\oplus
P),
    \label{recon}
\end{equation}
where, $I_{rec} \in \mathbb{R}^{h\times w \times 1}$ is the reconstructed image.
\vspace{-0.03in}
\subsection{Details in Implementation}
\vspace{-0.03in}
Our framework is implemented by PyTorch, and we use Polanalyser library\footnote{\url{https://github.com/elerac/polanalyser/wiki}} to process the polarized images. 
It is trained with $L1$ loss between $I_{rec}$ and $I_{gt}$ in an end-to-end manner.
Besides, the learning rate is $5\times 10^{-5}$, and it decays to half of the original for every $2\times 10^{4}$ iterations.

\begin{figure*}
	\centering
	\Huge
	\newcommand\widthface{0.4}
	\resizebox{1.0\linewidth}{!}{
	\begin{tabular}{ccccccc}
	    Input (Intensity) & SHIQ~\cite{fu-2021-multi-task}& Polar-HR~\cite{wang2022uformer} & Uformer~\cite{wang2022uformer} & Restormer~\cite{Zamir2021Restormer} & Ours & Ground Truth\\
	    \includegraphics[width=\widthface\textwidth]{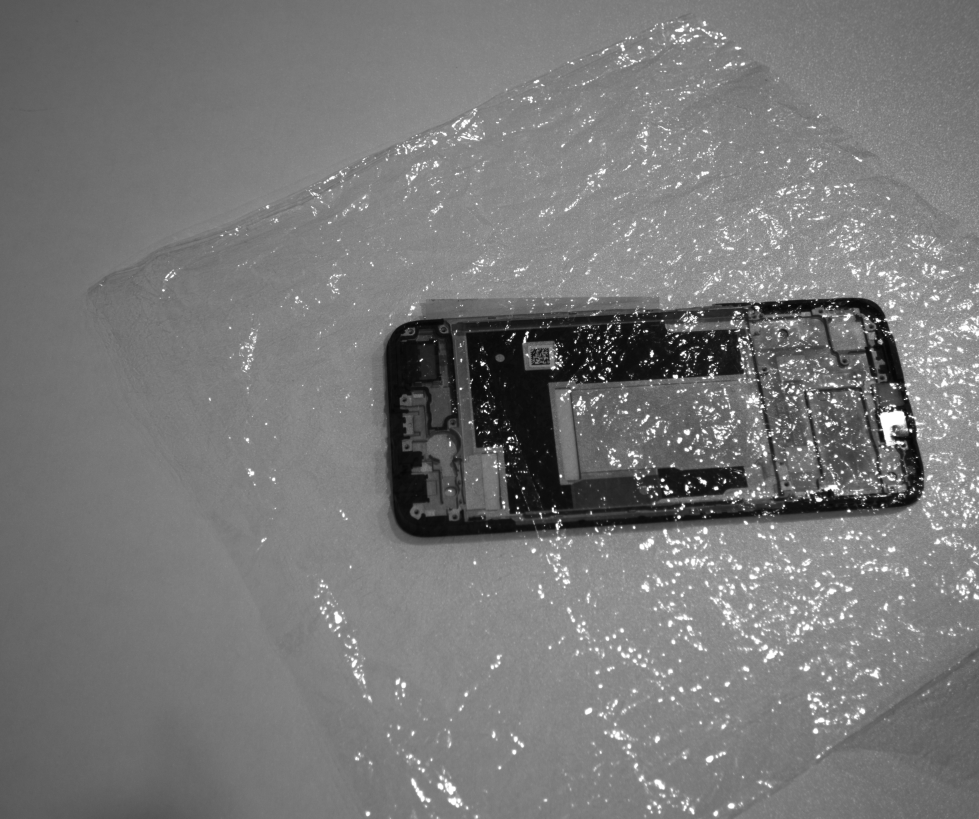} &
		\includegraphics[width=\widthface\textwidth]{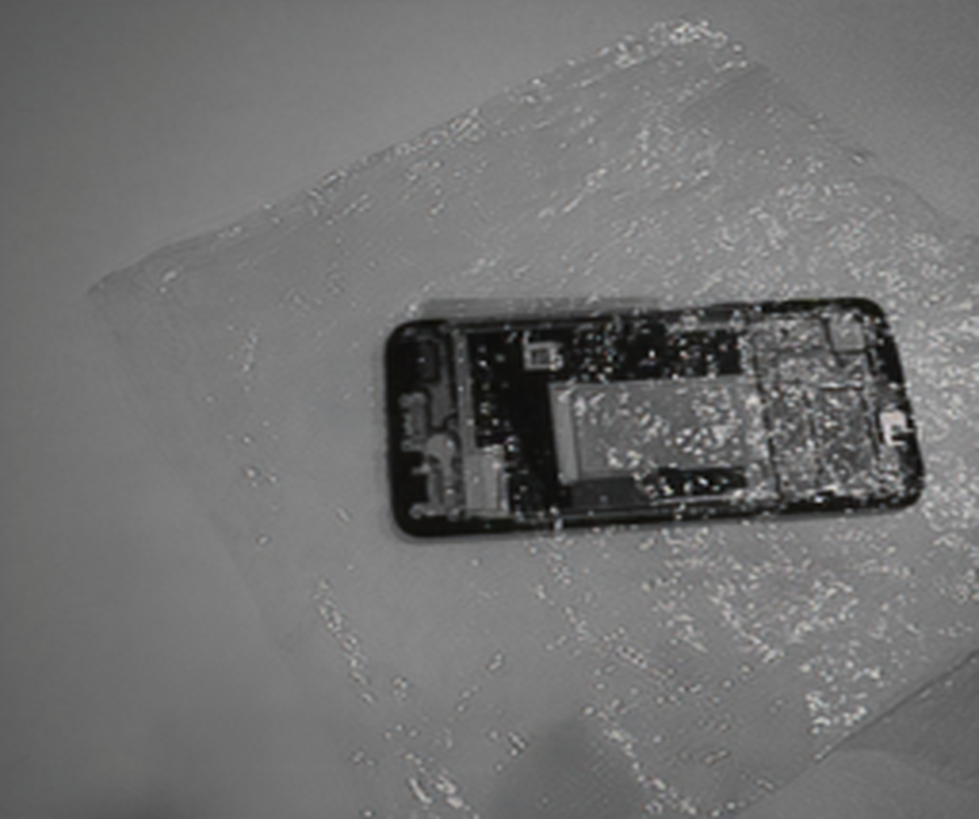} &
		\includegraphics[width=\widthface\textwidth]{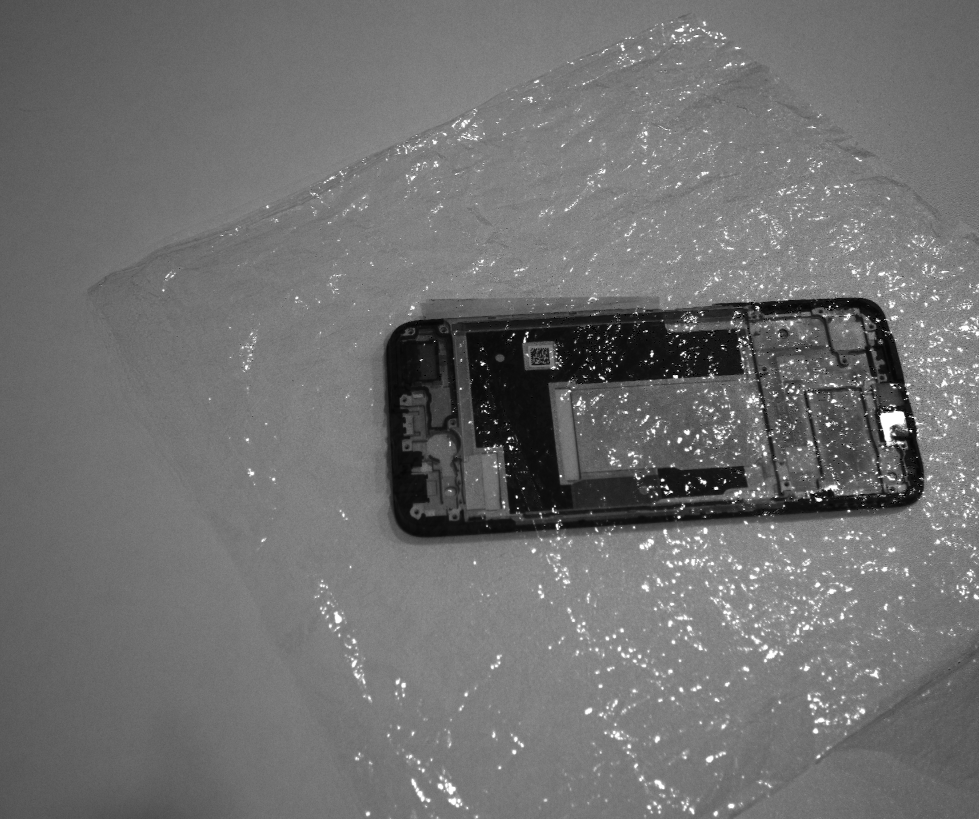} &	    
		\includegraphics[width=\widthface\textwidth]{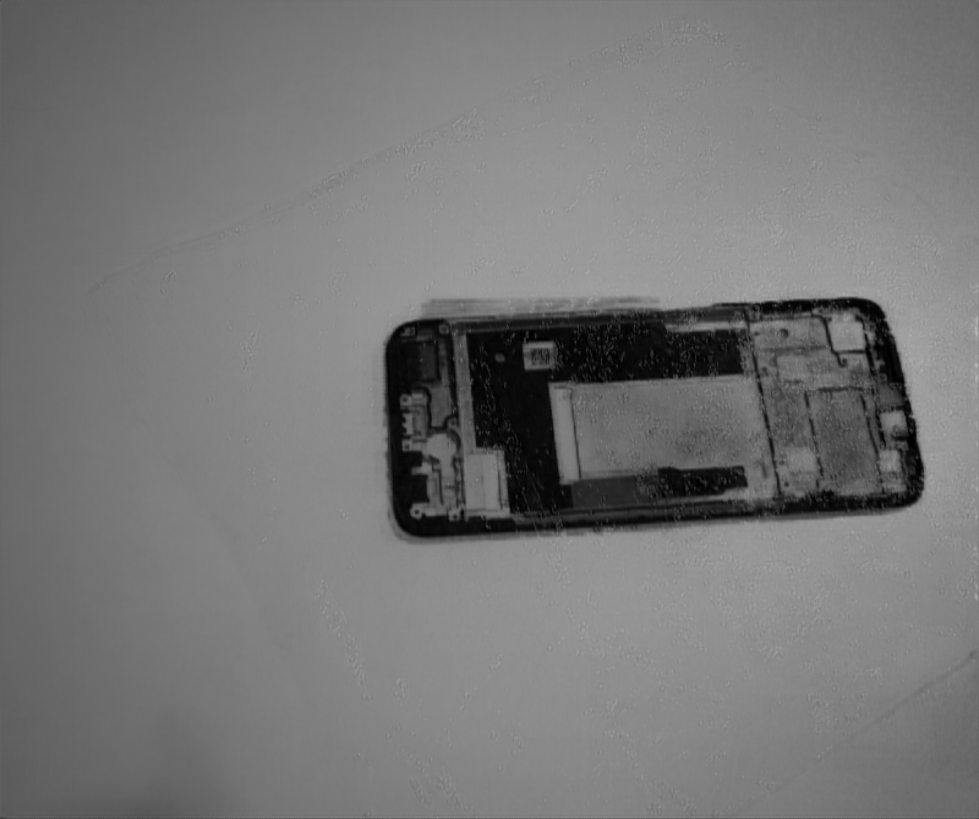} &
  		\includegraphics[width=\widthface\textwidth]{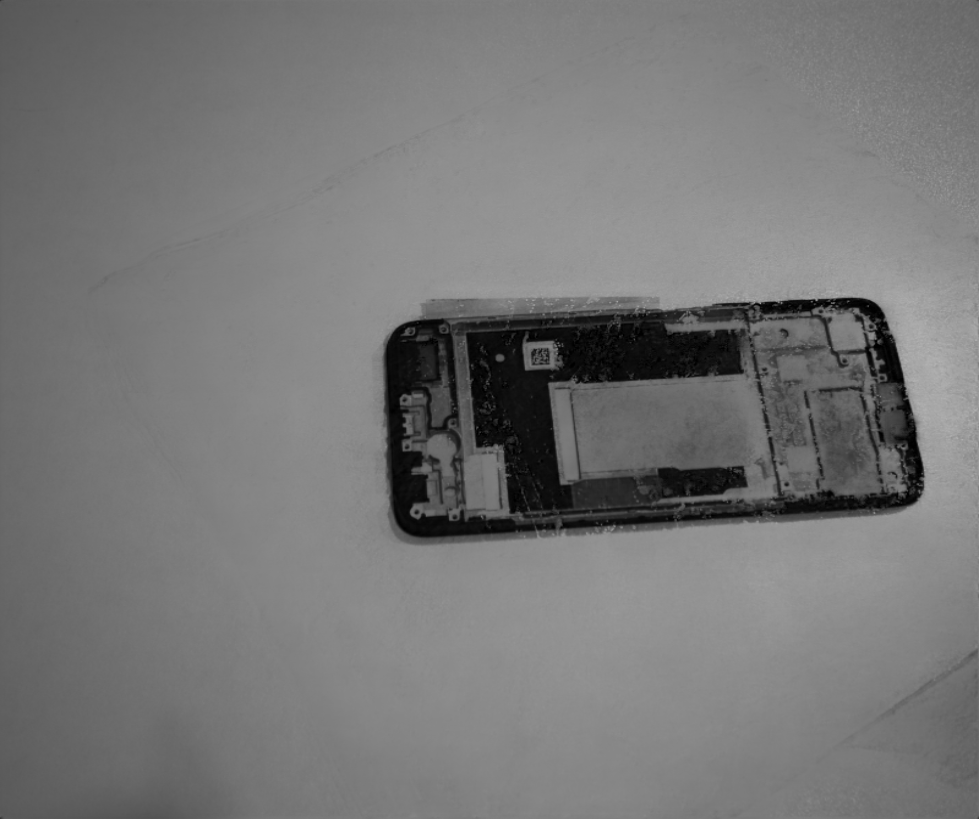} &
		\includegraphics[width=\widthface\textwidth]{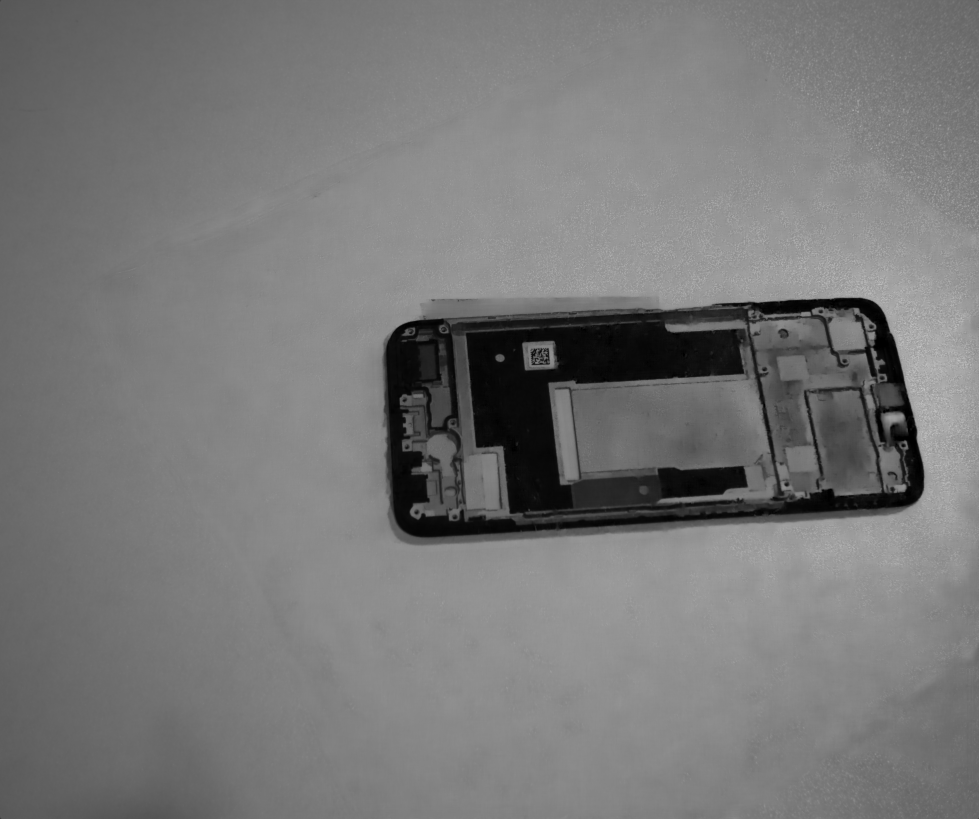} &
		\includegraphics[width=\widthface\textwidth]{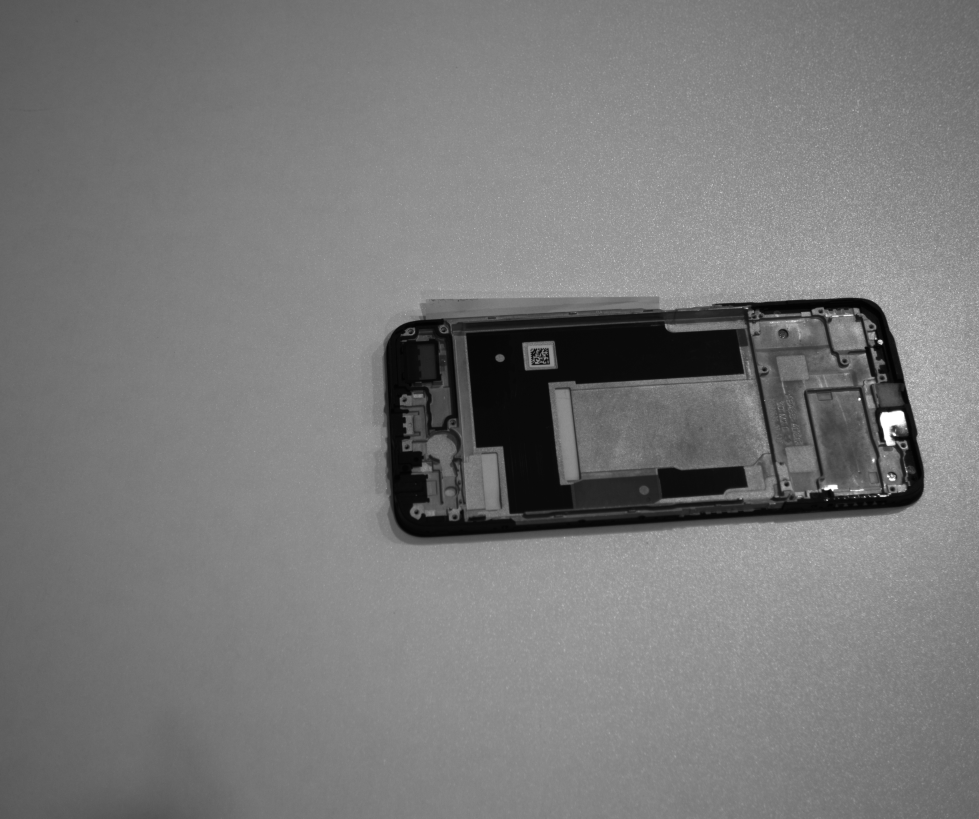} \\

  		\includegraphics[width=\widthface\textwidth]{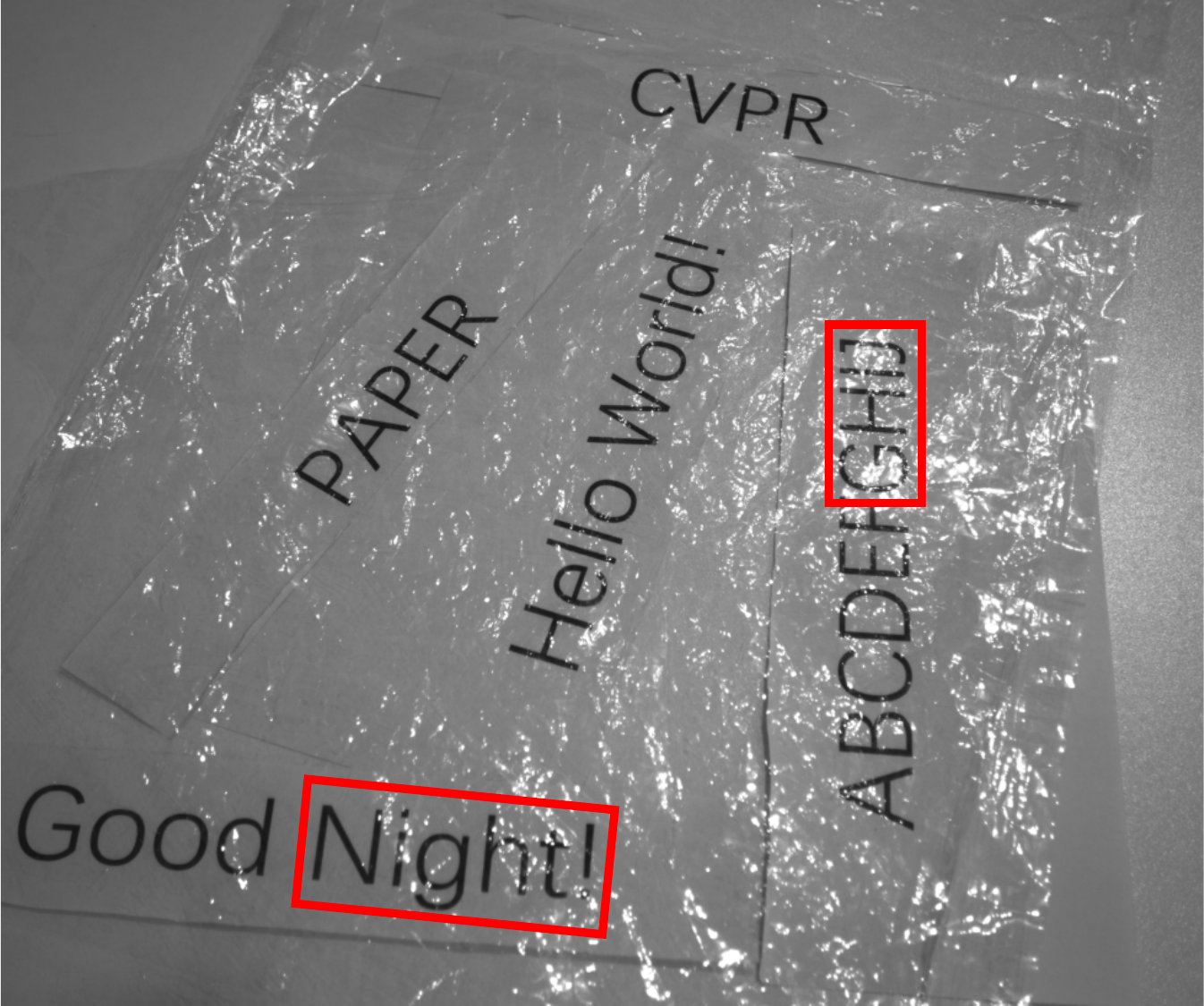} &
		\includegraphics[width=\widthface\textwidth]{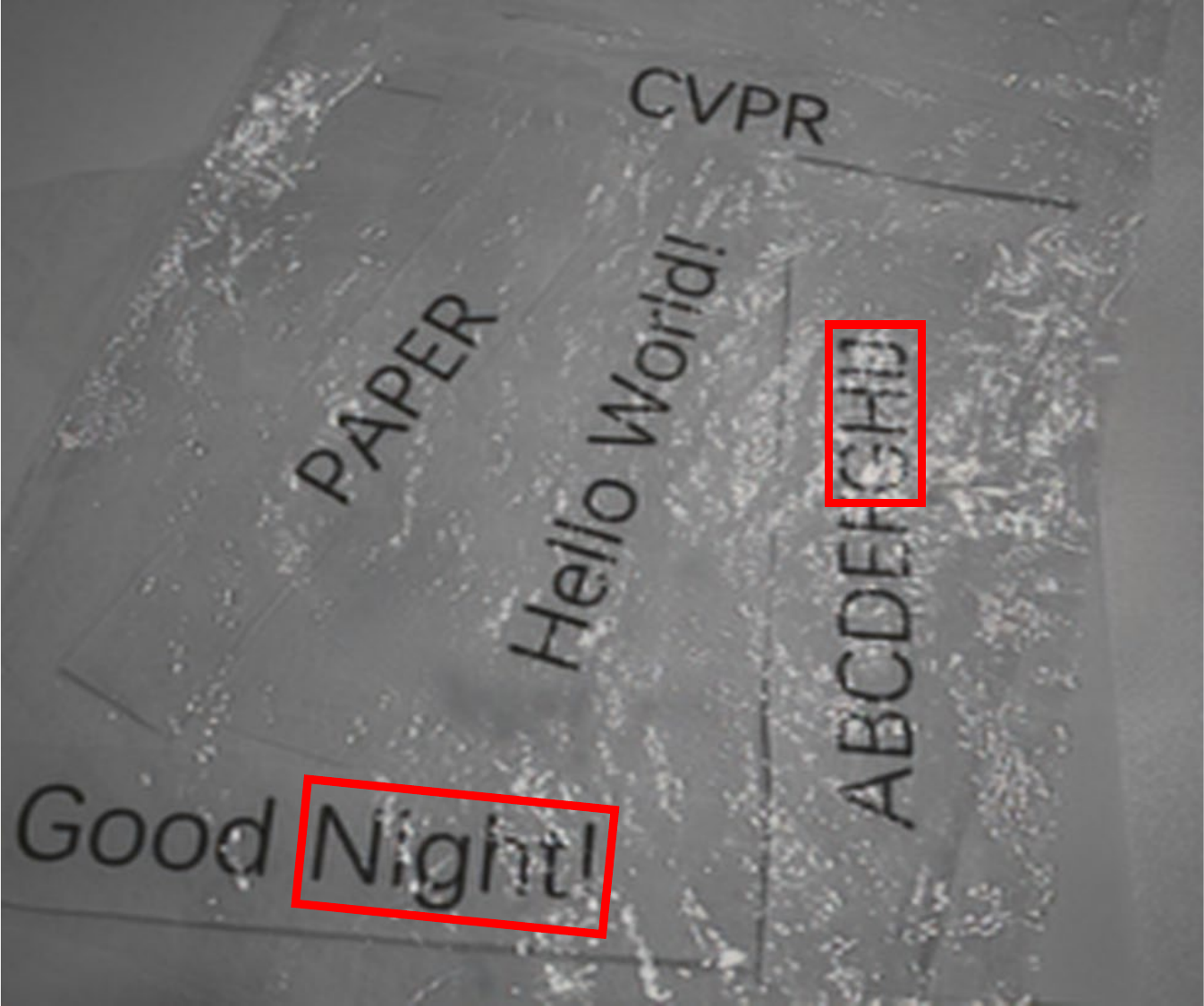} &    
		\includegraphics[width=\widthface\textwidth]{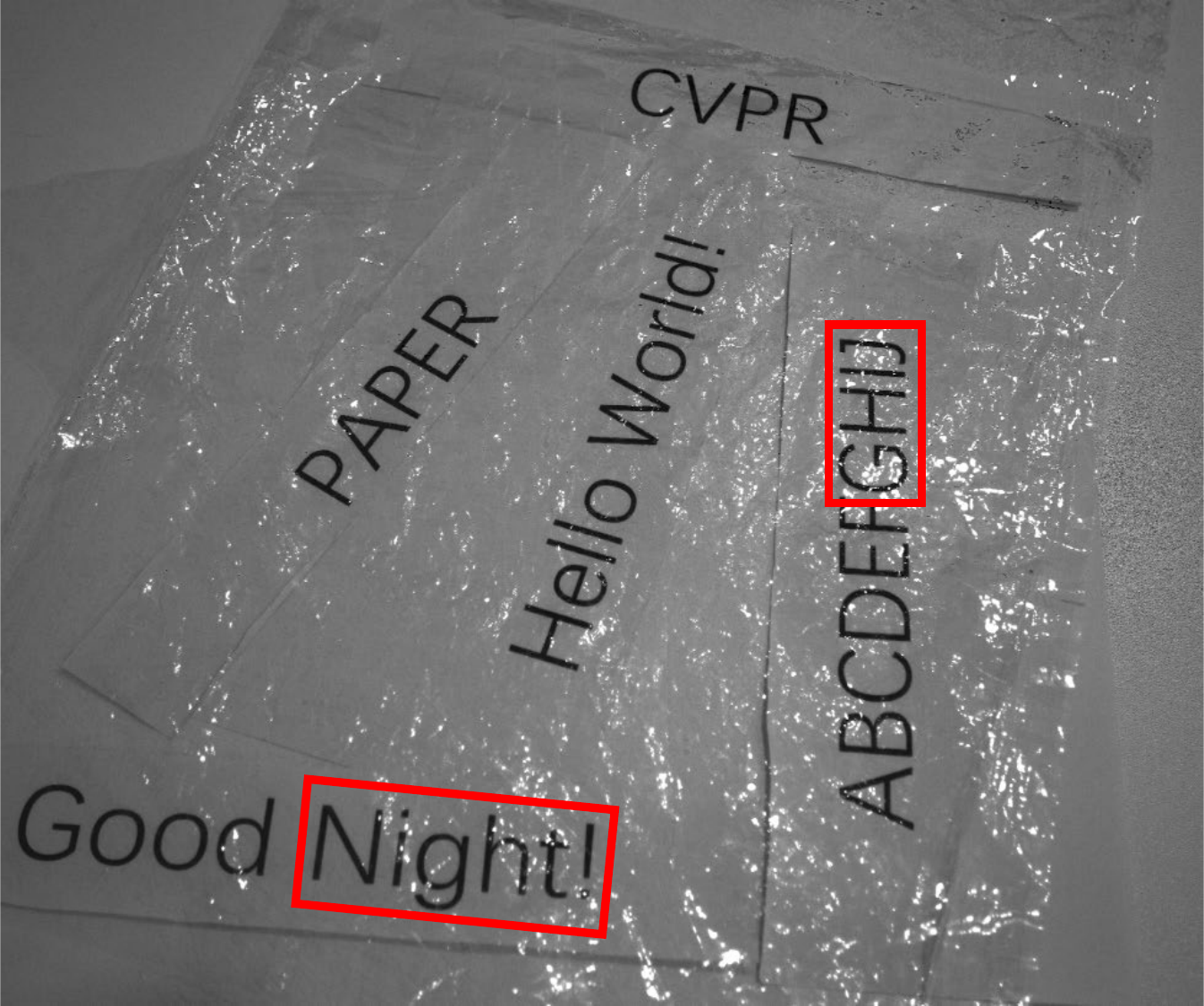} &
		\includegraphics[width=\widthface\textwidth]{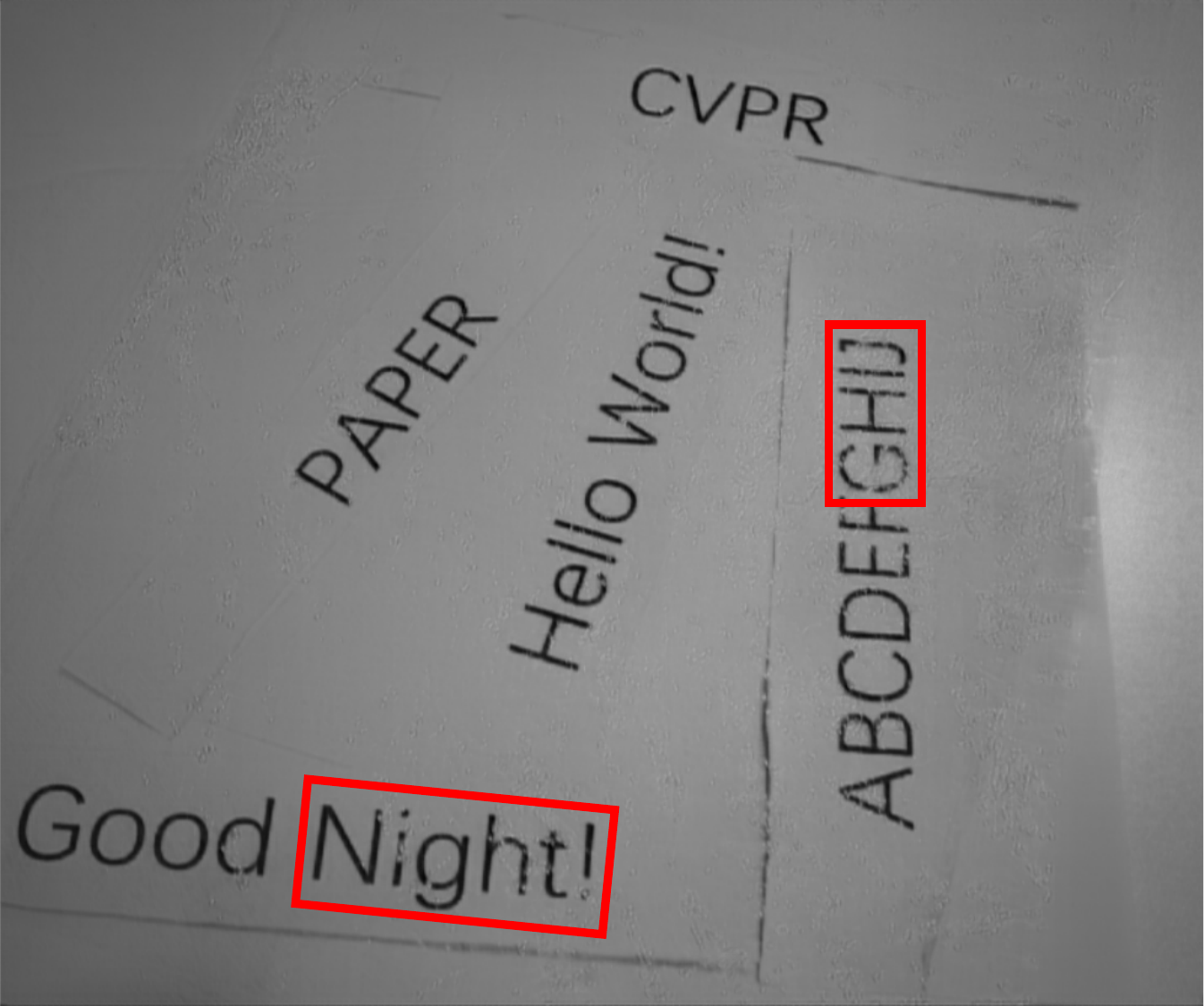} &
  		\includegraphics[width=\widthface\textwidth]{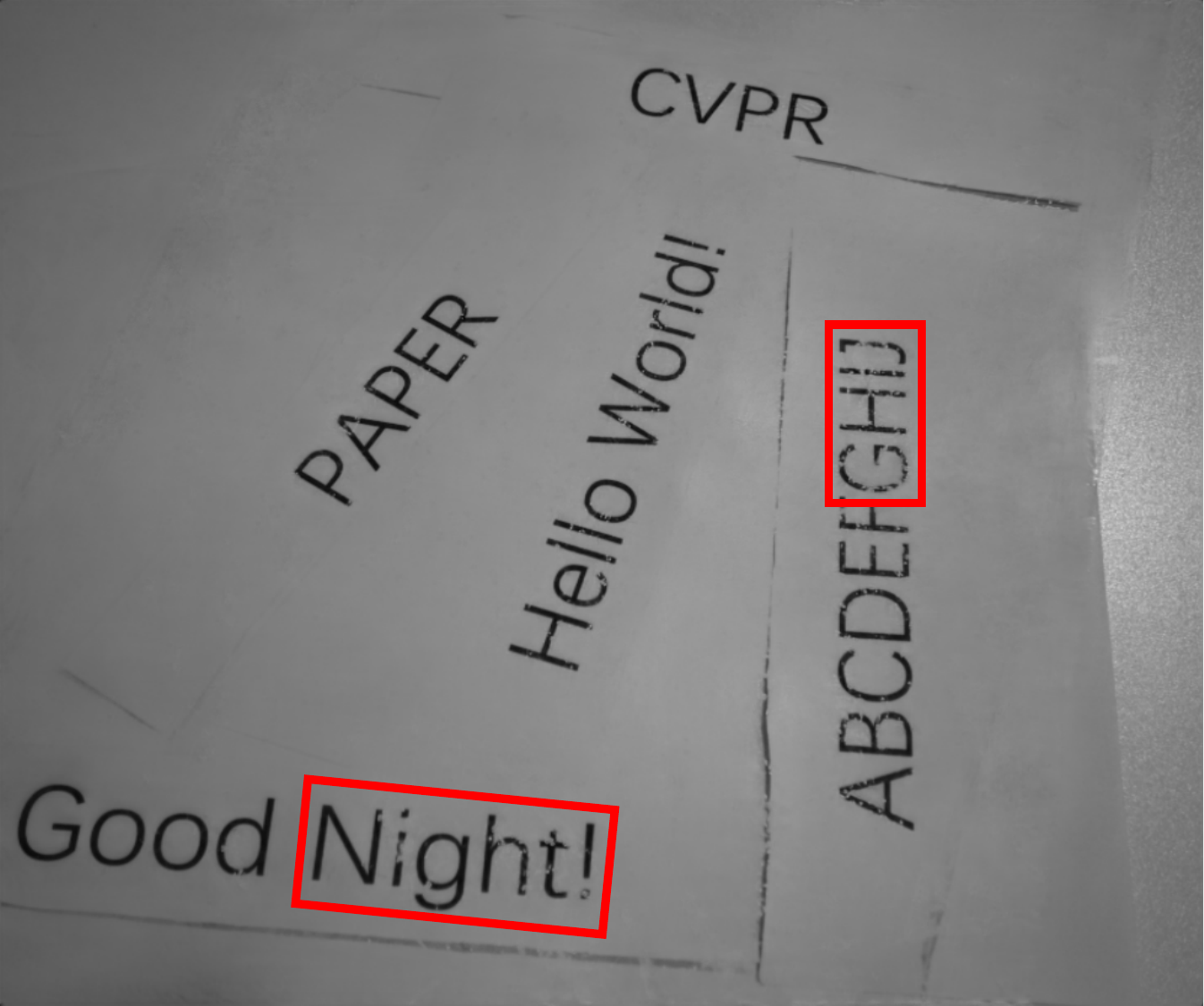} &
		\includegraphics[width=\widthface\textwidth]{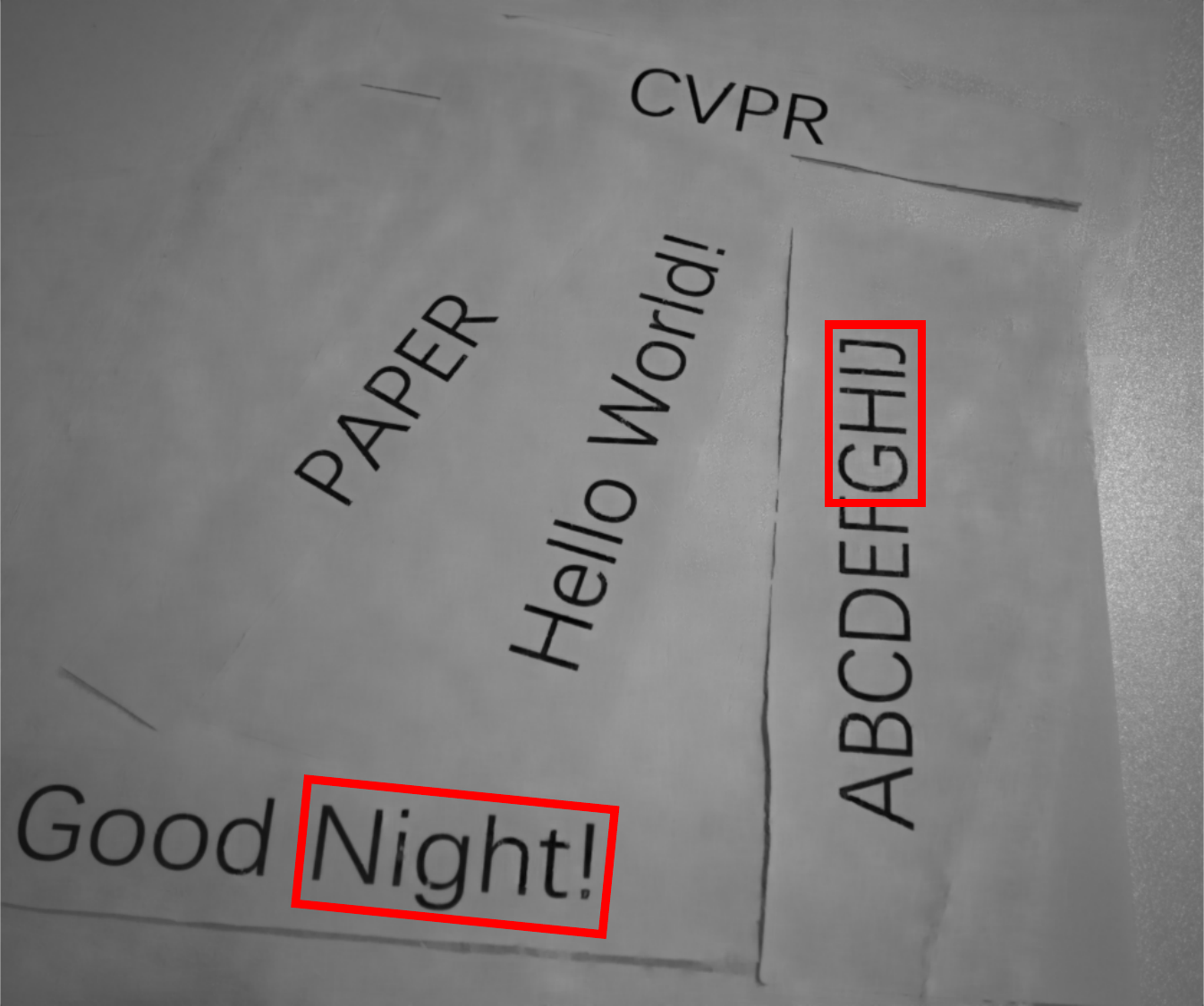} &
		\includegraphics[width=\widthface\textwidth]{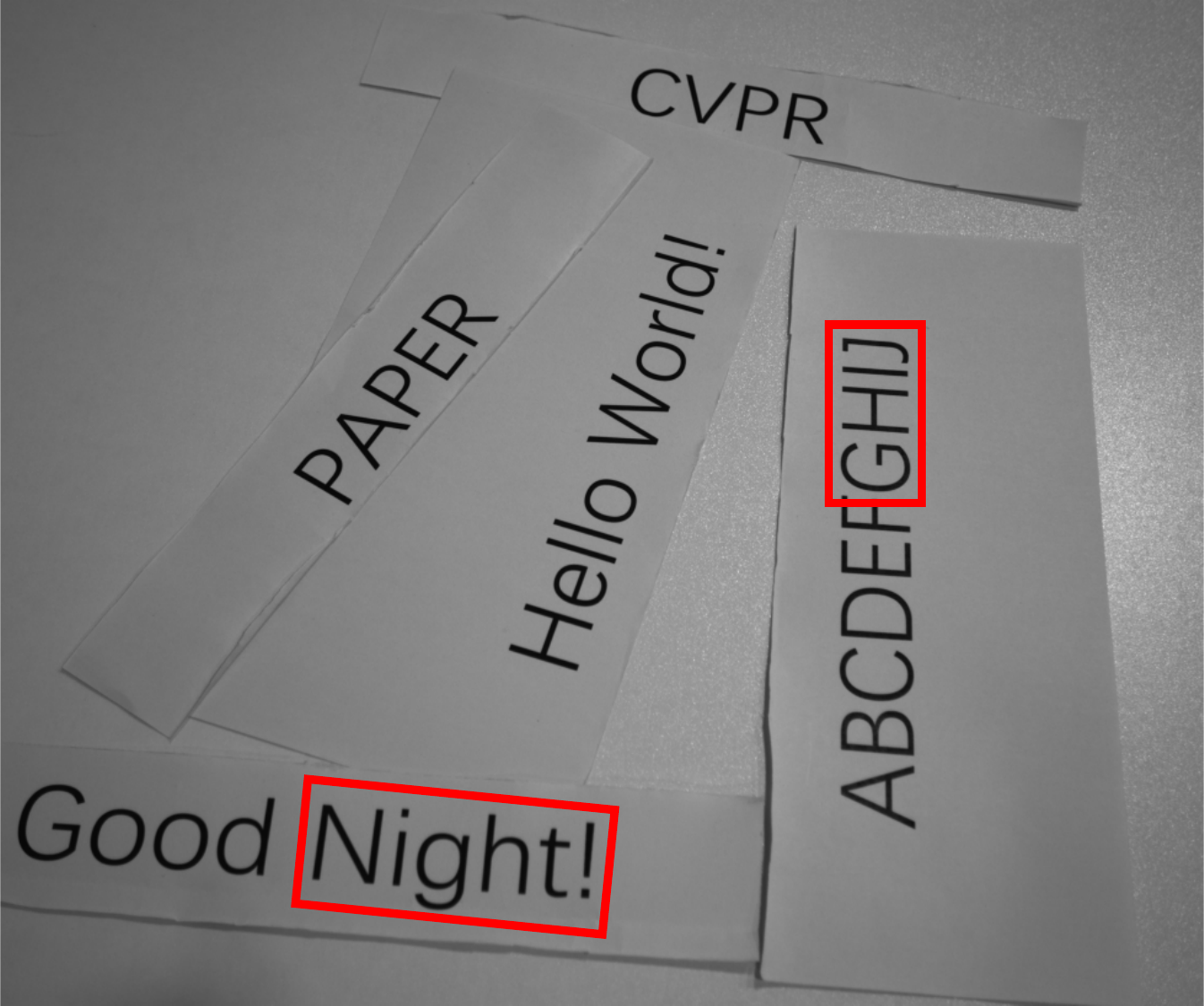} \\

  		\includegraphics[width=\widthface\textwidth]{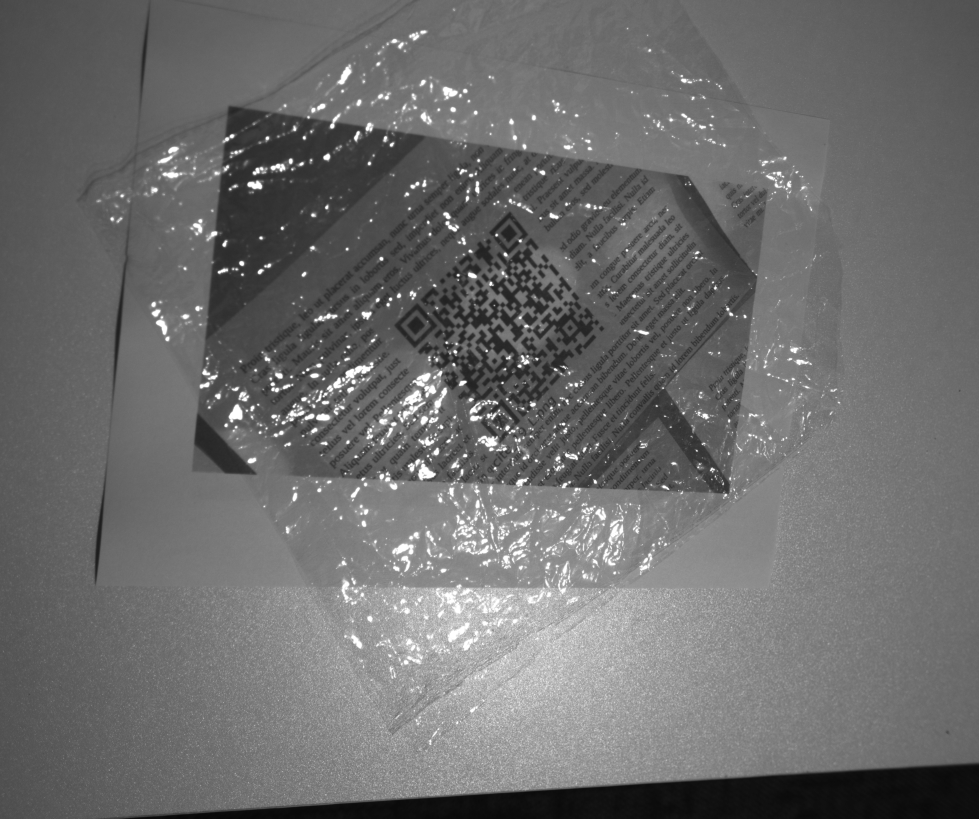} &
		\includegraphics[width=\widthface\textwidth]{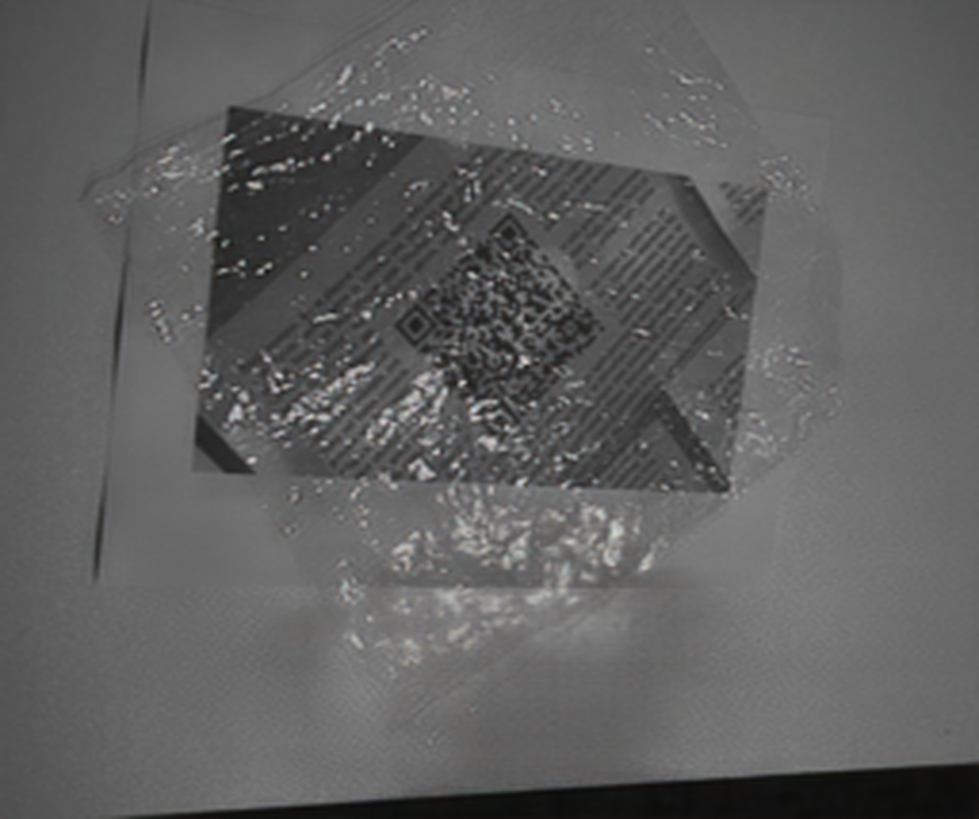} &    
		\includegraphics[width=\widthface\textwidth]{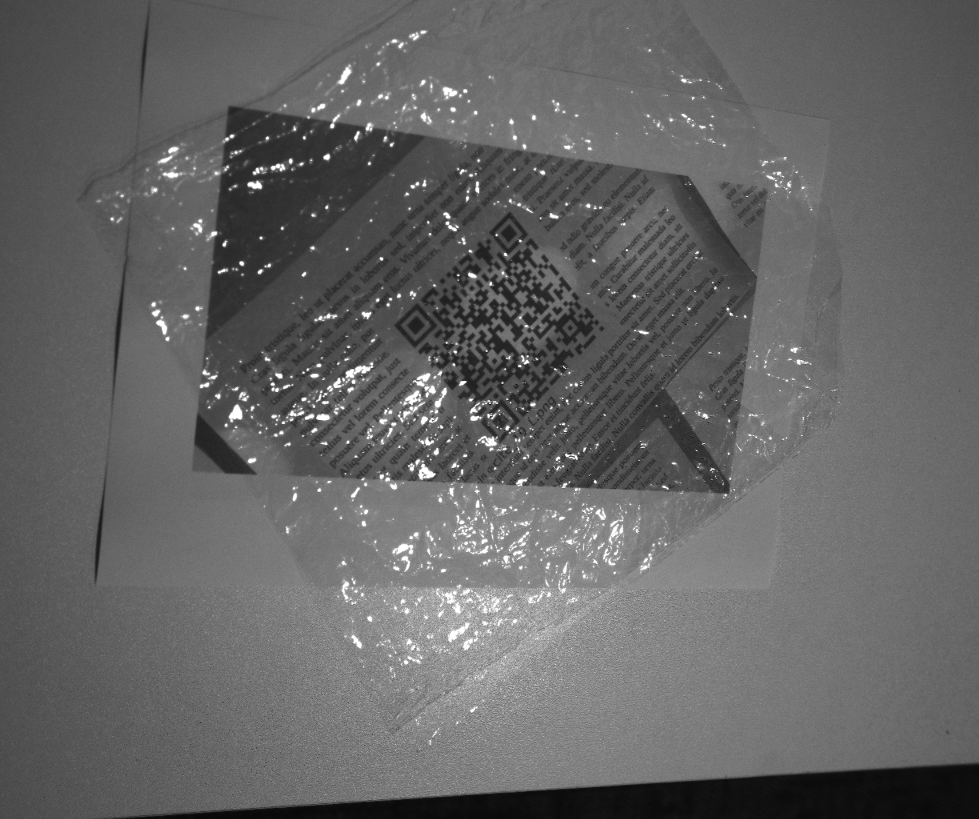} &
		\includegraphics[width=\widthface\textwidth]{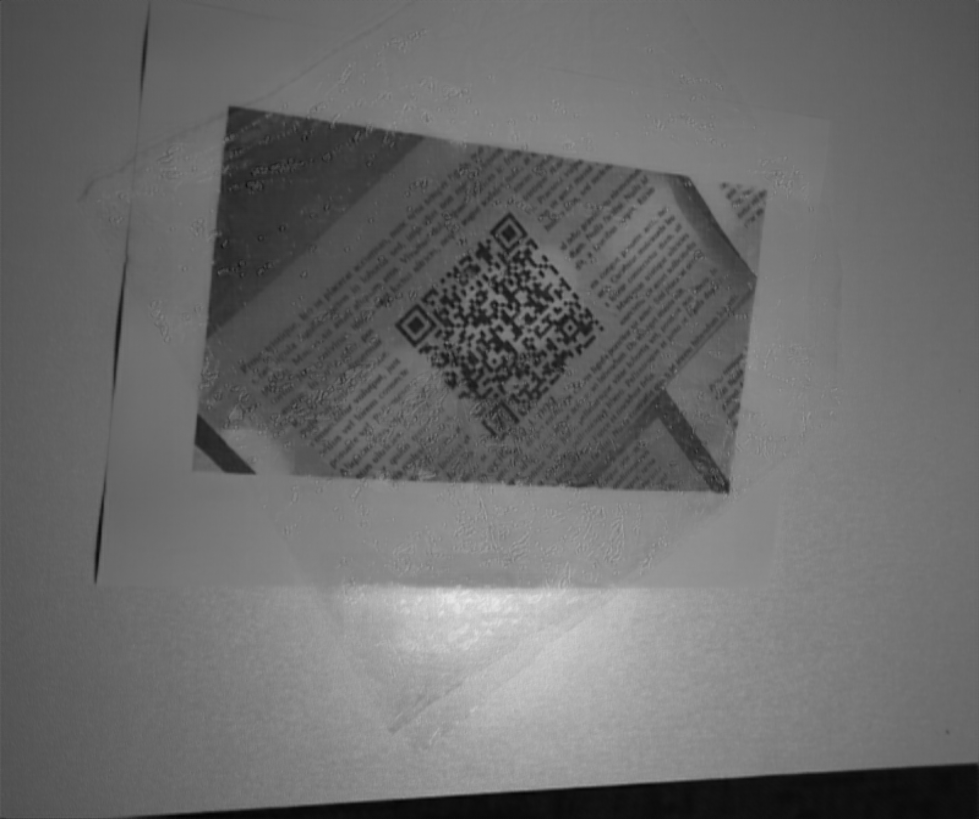} &
  		\includegraphics[width=\widthface\textwidth]{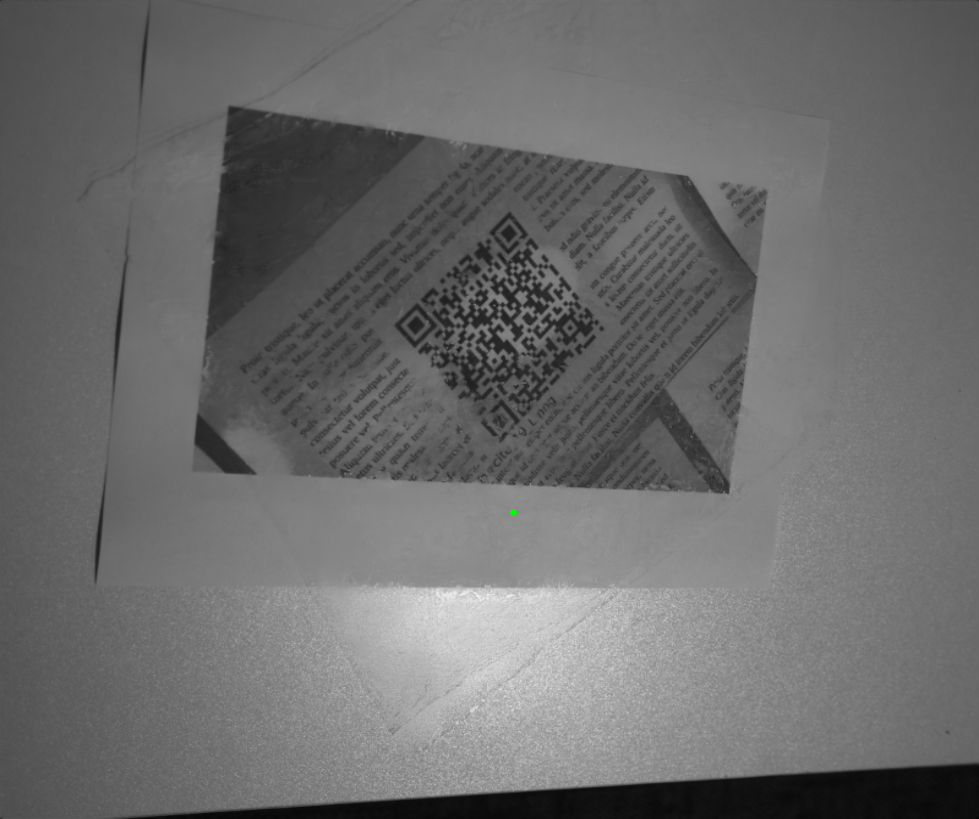} &
		\includegraphics[width=\widthface\textwidth]{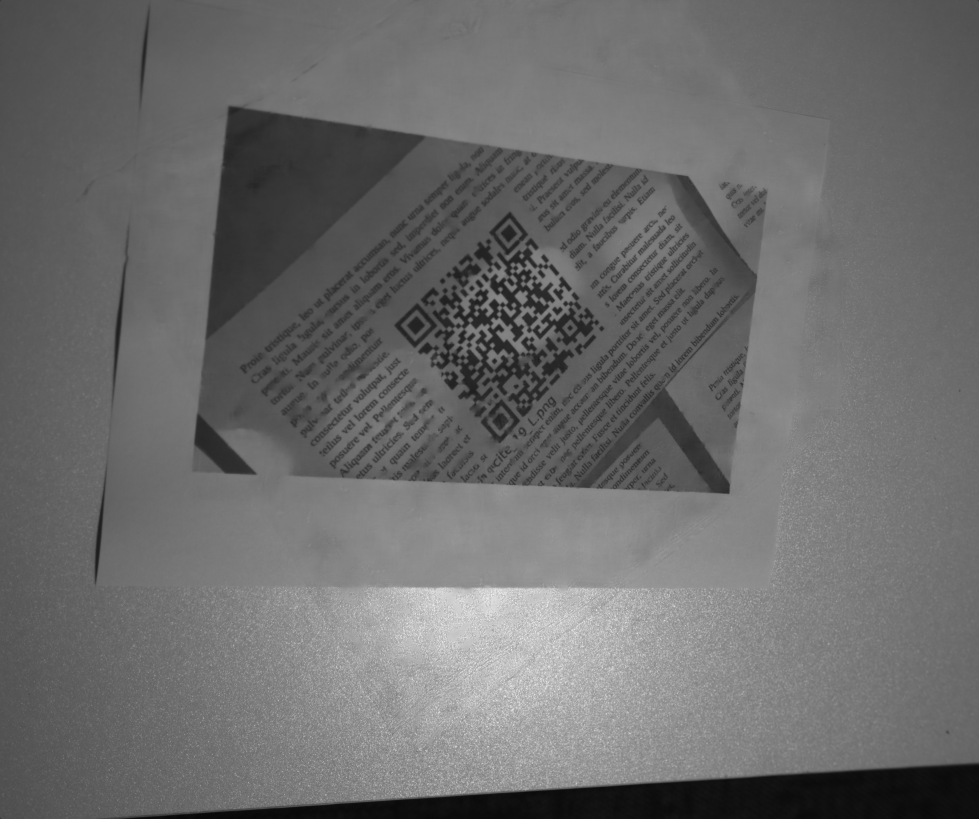} &
		\includegraphics[width=\widthface\textwidth]{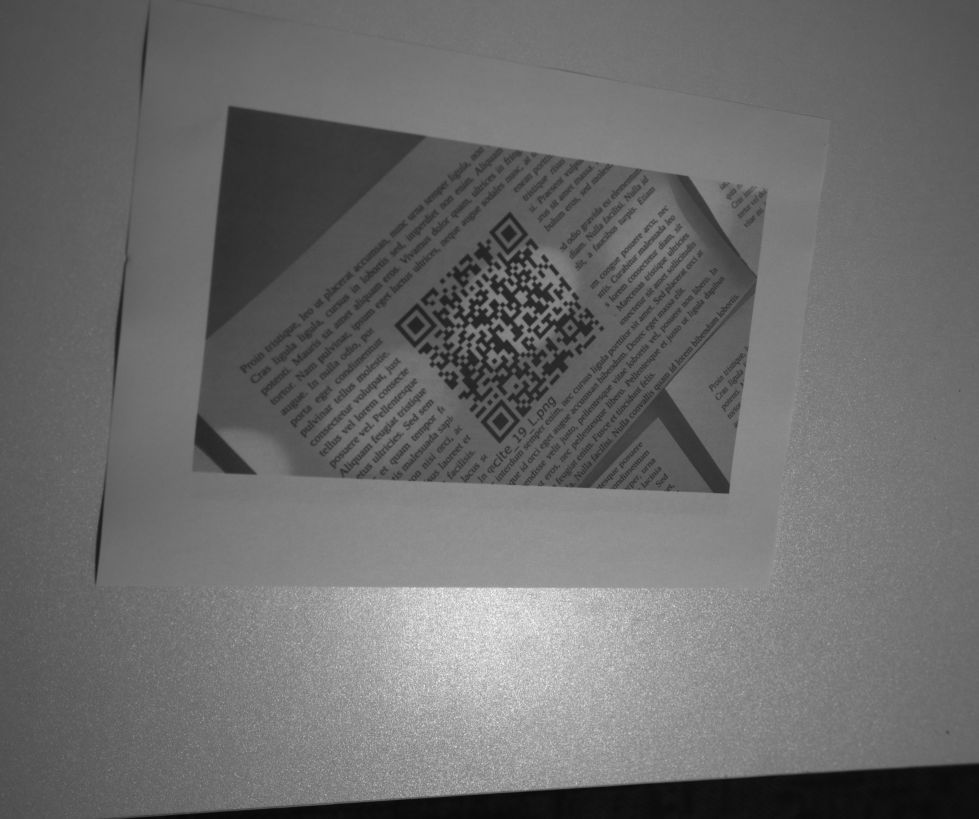}
	\end{tabular}}
	\vspace{-0.18in}
	\caption{Qualitative Evaluation. Compared with other baselines, our model can reconstruct more realistic details in highlight regions instead of fake artifacts. \color{red}Please zoom in for more details.}
	\vspace{-0.26in}
	\label{vis}
\end{figure*}

\vspace{-0.05in}
\section{Experiment}
\vspace{-0.05in}
\label{sec:ex}
In our experiment, we evaluate the performance of the proposed framework in three aspects. First, as an image reconstruction task, we conduct qualitative and quantitative evaluations with reconstruction-related metrics (\eg, PSNR, SSIM). Second, as an upstream task in the industry, two downstream scenarios (\eg, QR code reading, and Text OCR) are selected to evaluate the reliability of our proposed algorithm. Finally, we conduct ablation studies to analyze the roles of different components.
\vspace{-0.03in}
\subsection{Baselines}\label{visual}
\vspace{-0.05in}
There is no existing baseline for this new problem. Thus, we choose two general SOTA methods in image reconstruction, Uformer~\cite{wang2022uformer} and Restormer~\cite{Zamir2021Restormer} to evaluate the performance. Considering the connection between the film removal and highlight removal, we also compare with two SOTA polarization and unpolarization highlight removal baselines, Polar-HR~\cite{wen2021polarization} and SHIQ~\cite{fu-2021-multi-task}, respectively.

To make a fair comparison, we follow the same training setting in our framework and input four polarized images to reconstruct one image without wrinkled transparent film in Uformer~\cite{wang2022uformer}, Restormer~\cite{Zamir2021Restormer}. In Polar-HR~\cite{wen2021polarization}, it is a traditional training-free model and therefore shares the same input and output data as our approach. Besides, in SHIQ~\cite{fu-2021-multi-task}, since it cannot support polarized images, we convert four polarized images to one unpolarized image.
\vspace{-0.03in}
\subsection{Evaluation on Reconstruction Task}
\vspace{-0.05in}
\noindent\textbf{Quantitative Evaluation}
To measure the quantitative performance of the algorithm, we refer to commonly used metrics for Image Reconstruction: peak signal-to-noise ratio (PSNR) and structural similarity (SSIM). 
Also, to compare the overall performance of our dataset, we use 10-fold cross-validation during training and testing. This strategy ensures the evaluation's reliability and further proves the proposed algorithm's robustness.

The quantitative results are shown in Table~\ref{cross}. In the last column of Table~\ref{cross}, $\mu$ is the mean value, and $\sigma$ is the variance of the 10-fold performance. These experimental results show the average PSNR is 36.48 and the average SSIM is 0.9824, and the results demonstrate the high quality of our reconstructed images.
In addition, the variance of PSNR and SSIM are 0.57 and $1.23 \times 10^{-5}$, respectively, and this proves the proposed algorithm is stable and robust.

\noindent\textbf{Qualitative Evaluation} To evaluate the reconstruction performance of the algorithm in a qualitative way, we show some visual results. Our results are shown in Fig.~\ref{vis}. Although Polar-HR~\cite{wen2021polarization} and SHIQ~\cite{fu-2021-multi-task} can remove some of the highlights, their models cannot model the film correctly and, therefore, cannot remove the film itself. Besides, it can be observed that our algorithm can reconstruct regions where text or QR codes are corrupted by highlights through polarization information, while Uformer~\cite{wang2022uformer} and Restormer~\cite{Zamir2021Restormer} produce fake artifacts in these regions.
\begin{table}
  \centering
    \resizebox{1.0\linewidth}{!}{
    \begin{tabular}{l|ccc}
    \toprule
     & SHIQ~\cite{fu-2021-multi-task} &  Polar-HR~\cite{wen2021polarization} & Uformer~\cite{wang2022uformer} \\
     \midrule 
     Read Number  & 7    & 17  & 18 \\
     Read Rate  & 8.05\% & 19.54\% & 20.69\%\\
    \midrule 
     & Restormer~\cite{Zamir2021Restormer}  & Ours  & Ground Truth \\
     \midrule 
    Read Number  & 24    & \textbf{63}    & 87 \\
    Read Rate & 27.58\%  & \textbf{72.41\%} & 100\% \\
    \bottomrule
    \end{tabular}}
    \vspace{-0.15in}
  \caption{QR code reading rate. Compared with other baselines, our approach can improve the performance of QR codes scanner in the industrial environment.}
    \vspace{-0.33in}
  \label{tQR}
\end{table}%

\vspace{-0.03in}
\subsection{Evaluation on Downstream Applications}
\vspace{-0.03in}
In industrial environments, it is common to use transparent films to cover products, 
which may negatively impact the robustness of downstream algorithms.
To evaluate the effectiveness of our solution in such settings, 
we conduct two tests on two downstream tasks:
QR Code Reading and Text Optical Character Recognition (Text OCR). 
These tasks are particularly relevant as they require access to raw information, which provides a rigorous test of the effectiveness of our algorithms in industrial scenarios.

\begin{figure}
\vspace{-0.08in}
	\centering
	\Huge
	\newcommand\widthface{0.45}
	\resizebox{1.0\linewidth}{!}{
	\begin{tabular}{ccc}
	    Input (Intensity) & SHIQ~\cite{fu-2021-multi-task} & Polar-HR~\cite{wen2021polarization}\\
	    \includegraphics[width=\widthface\textwidth]{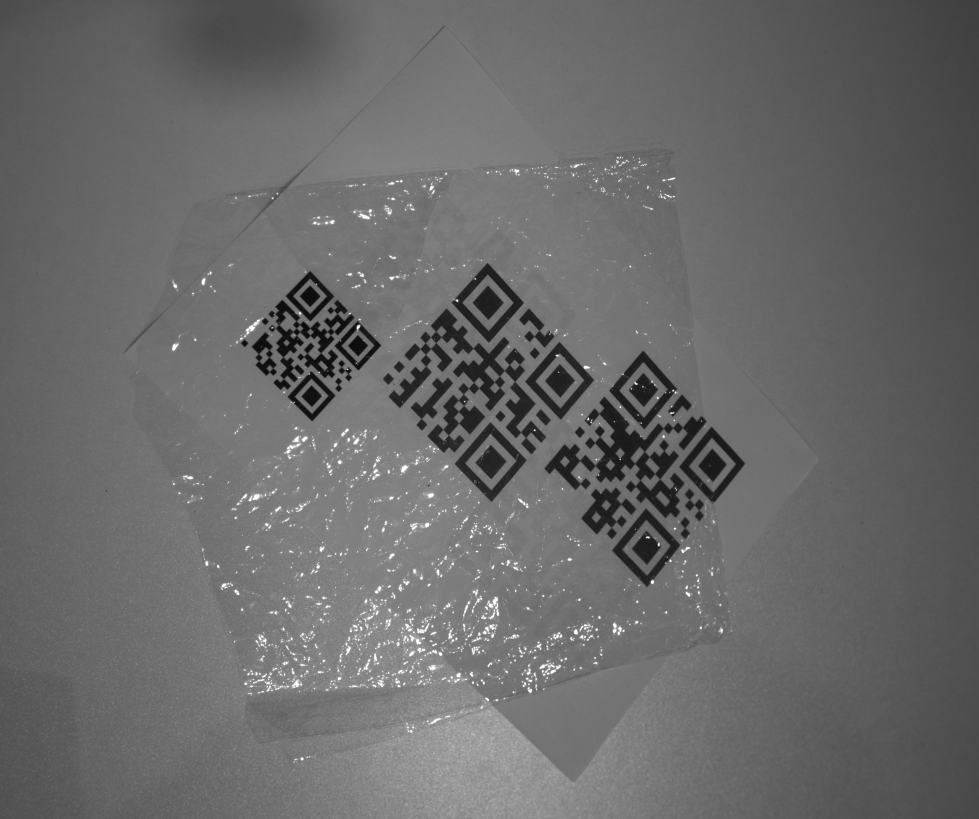} &
     	\includegraphics[width=\widthface\textwidth]{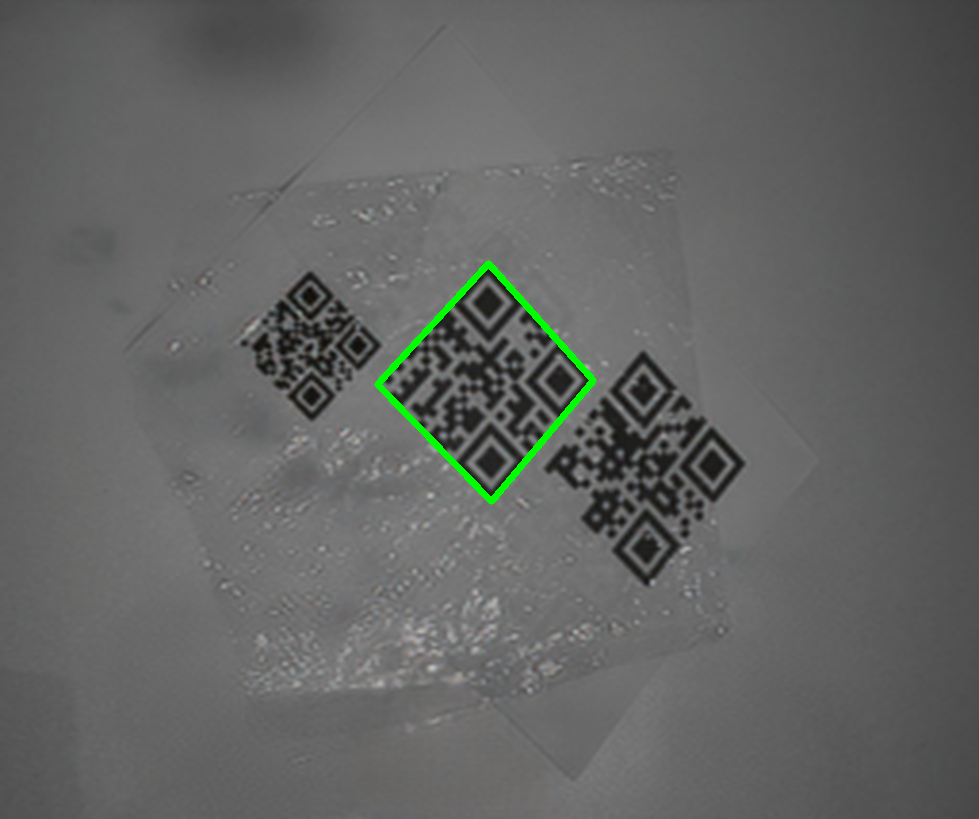} &
		\includegraphics[width=\widthface\textwidth]{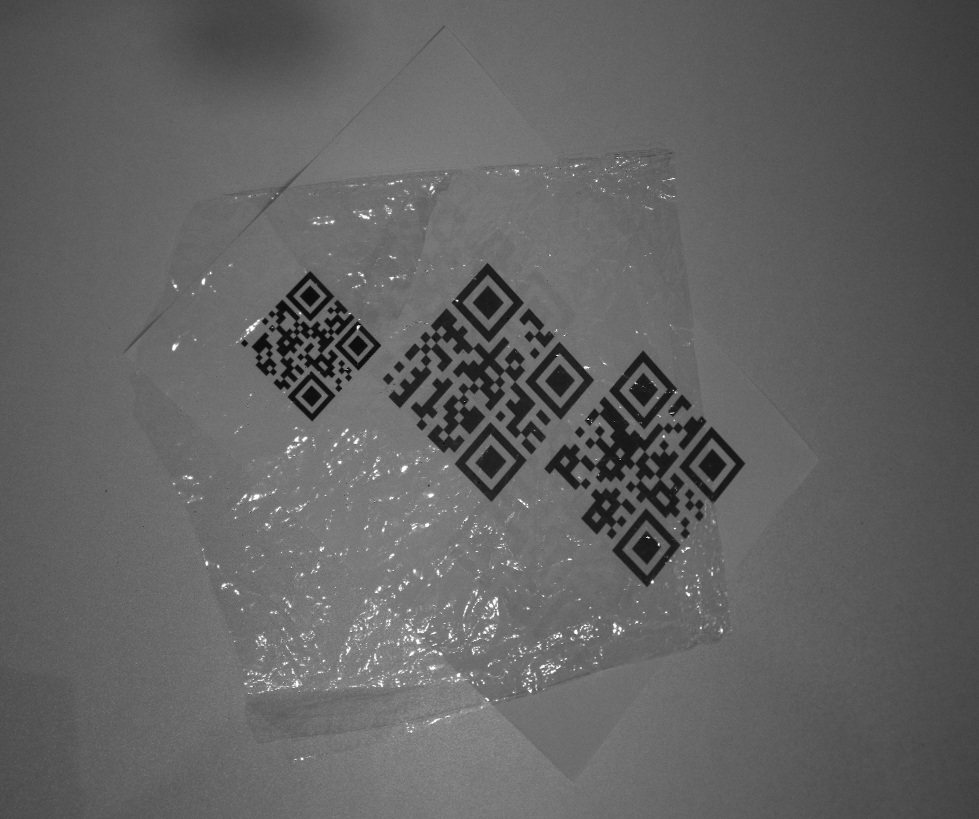} \\
             Uformer~\cite{wang2022uformer} & Restormer~\cite{Zamir2021Restormer} & Ours  \\
	    \includegraphics[width=\widthface\textwidth]{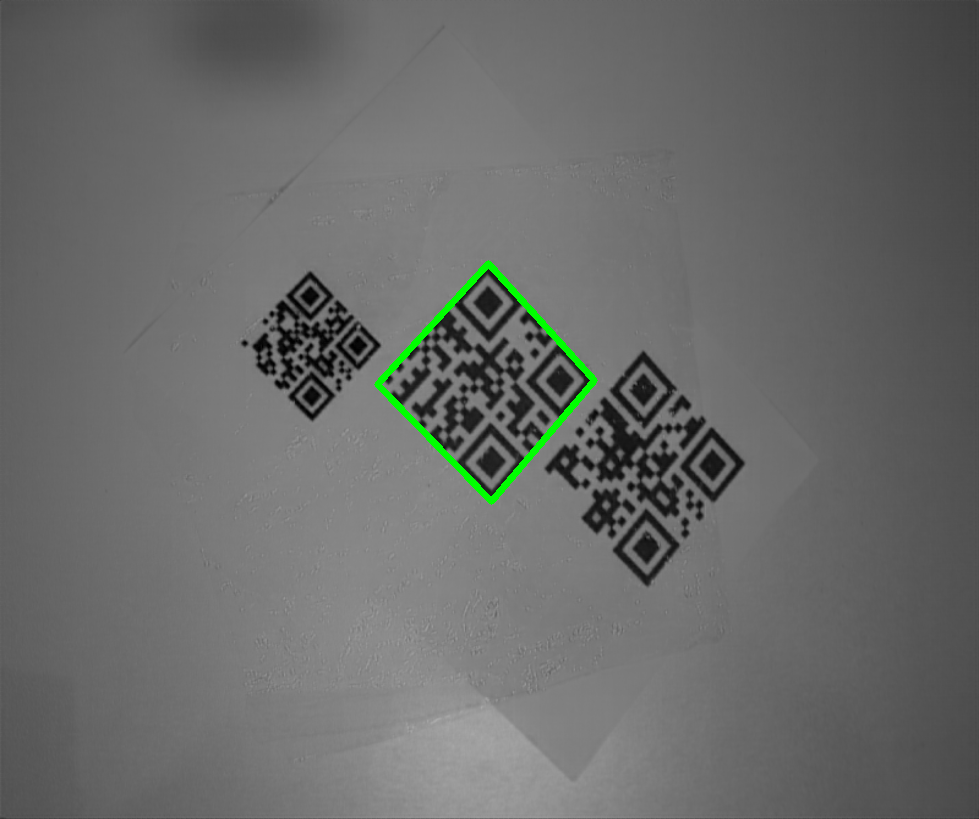} &
     	\includegraphics[width=\widthface\textwidth]{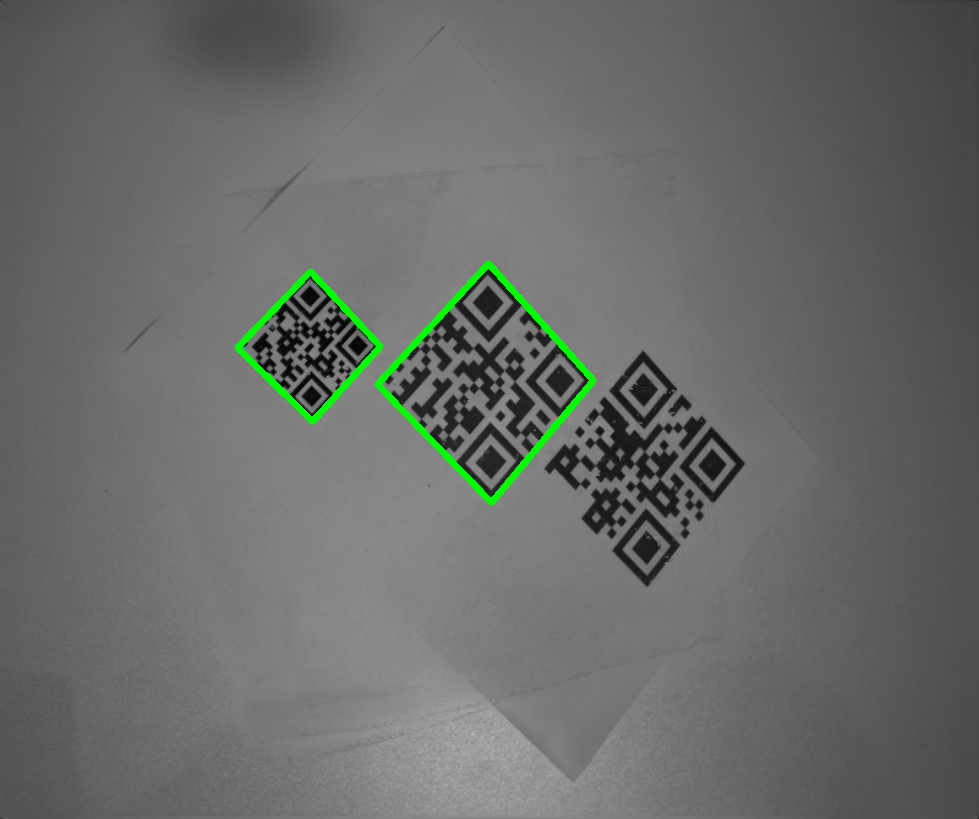} &
		\includegraphics[width=\widthface\textwidth]{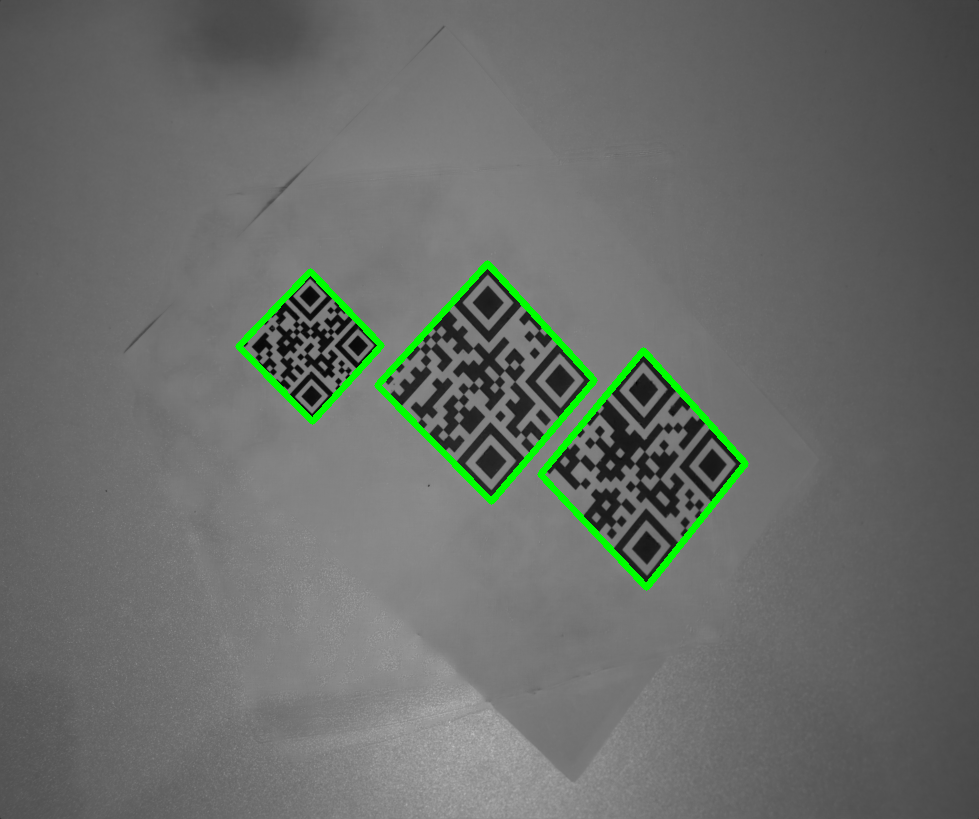}
	\end{tabular}}
        \vspace{-0.19in}
	\caption{Performance of QR code reading in the industry. In the original image, the QR code scanner fails to detect QR code information. Polar-HR~\cite{wen2021polarization} and SHIQ~\cite{fu-2021-multi-task} can only eliminate the specular reflection on the QR code, and Uformer~\cite{wang2022uformer} and Restormer~\cite{Zamir2021Restormer} can only generate false artifacts in highlight regions. Our method can help industrial QR code scanners achieve a higher performance.}
	\vspace{-0.19in}
	\label{tQRv}
\end{figure}

\noindent\textbf{QR Code Reading} The read rate is a critical performance metric for manufacturing pipelines in industrial systems. In this experiment, we compare our method with other baselines in Table~\ref{tQR}. Our upstream algorithm leads to a significant improvement in the performance of the QR code reading. The qualitative results in Fig.~\ref{tQRv} demonstrate that our algorithm not only removes the film but also significantly recovers the QR code information under the film.

\begin{figure}
	\centering
	\Huge
	\newcommand\widthface{0.45}
	\resizebox{0.95\linewidth}{!}{
	\begin{tabular}{ccc}
	    Input (Intensity) & SHIQ~\cite{fu-2021-multi-task} & Polar-HR~\cite{wen2021polarization}\\
	    \includegraphics[width=\widthface\textwidth]{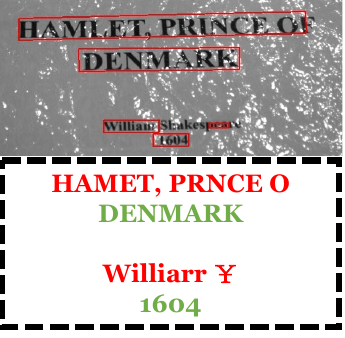} &
	    \includegraphics[width=\widthface\textwidth]{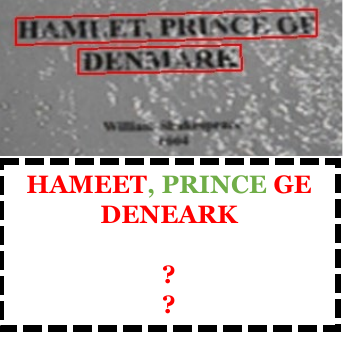} &
		\includegraphics[width=\widthface\textwidth]{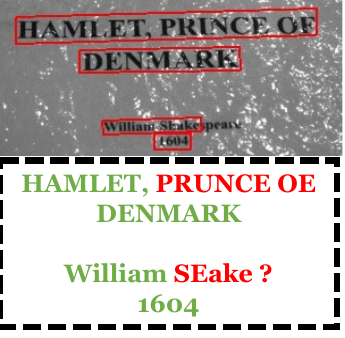} \\
            Uformer~\cite{wang2022uformer} & Restormer~\cite{Zamir2021Restormer} & Ours  \\
	    \includegraphics[width=\widthface\textwidth]{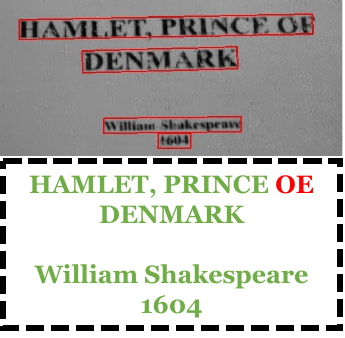} &
     	\includegraphics[width=\widthface\textwidth]{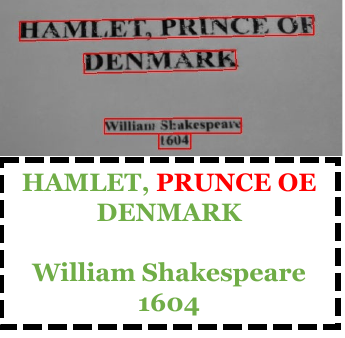} &
		\includegraphics[width=\widthface\textwidth]{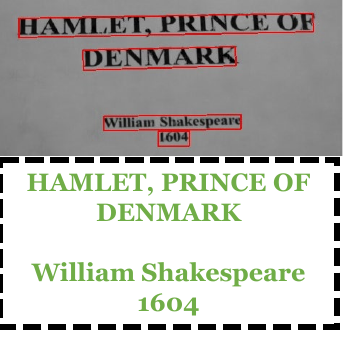}
	\end{tabular}}
	\vspace{-0.19in}
	\caption{Performance of text OCR in the industry. Compared to other baselines, our method can reconstruct more original text information for recognition.}
	\vspace{-0.14in}
	\label{text}
\end{figure}

\noindent\textbf{Text OCR} Text OCR is also an important downstream industrial task. We compared our algorithm with other baselines, as shown in Fig.~\ref{text}. Our method can restore more hidden text information accurately.

\vspace{-0.07in}
\subsection{Ablation Study}
\vspace{-0.07in}


\begin{figure}[t]
  \centering
  \vspace{-0.05in}
  \newcommand\widthface{1.0}
  \resizebox{0.96\linewidth}{!}{
  \includegraphics[width=\widthface\textwidth]{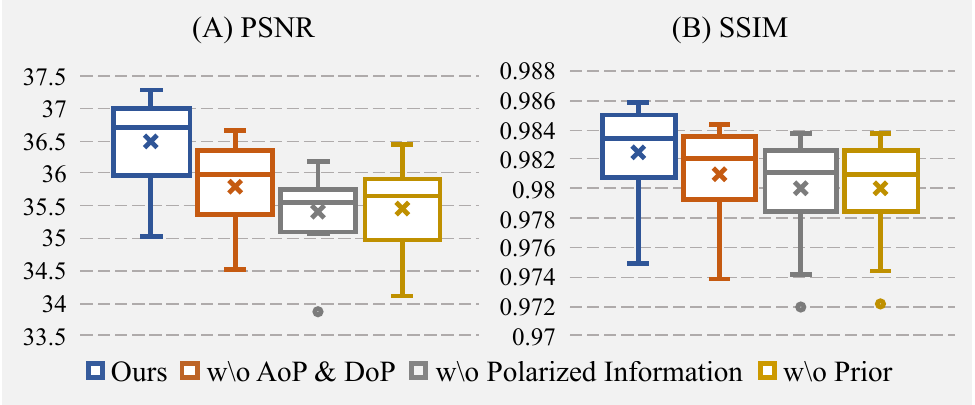}  
  }
  \vspace{-0.13in}
  \caption{PSNR (A) and SSIM (B) in ablation study. We use two boxplots to describe the performance after $t$-fold cross-validation. The results show that our dataset, the extracted prior, the AoP, and the DoP contribute to the framework performance improvement.}
  \vspace{-0.28in}
  \label{fig:abpsnr}
\end{figure}

We conduct ablation studies, which mainly validate the effectiveness of our proposed polarization dataset and one crucial component in our framework. To ensure that the experimental settings are consistent, we perform 10-fold cross-validation for the ablation experiments as well and draw two boxplots for further analysis.

\begin{figure}
	\centering
	\LARGE
	\newcommand\widthface{1}
	\resizebox{0.95\linewidth}{!}{
	\begin{tabular}{c}
		\includegraphics[width=\widthface\textwidth]{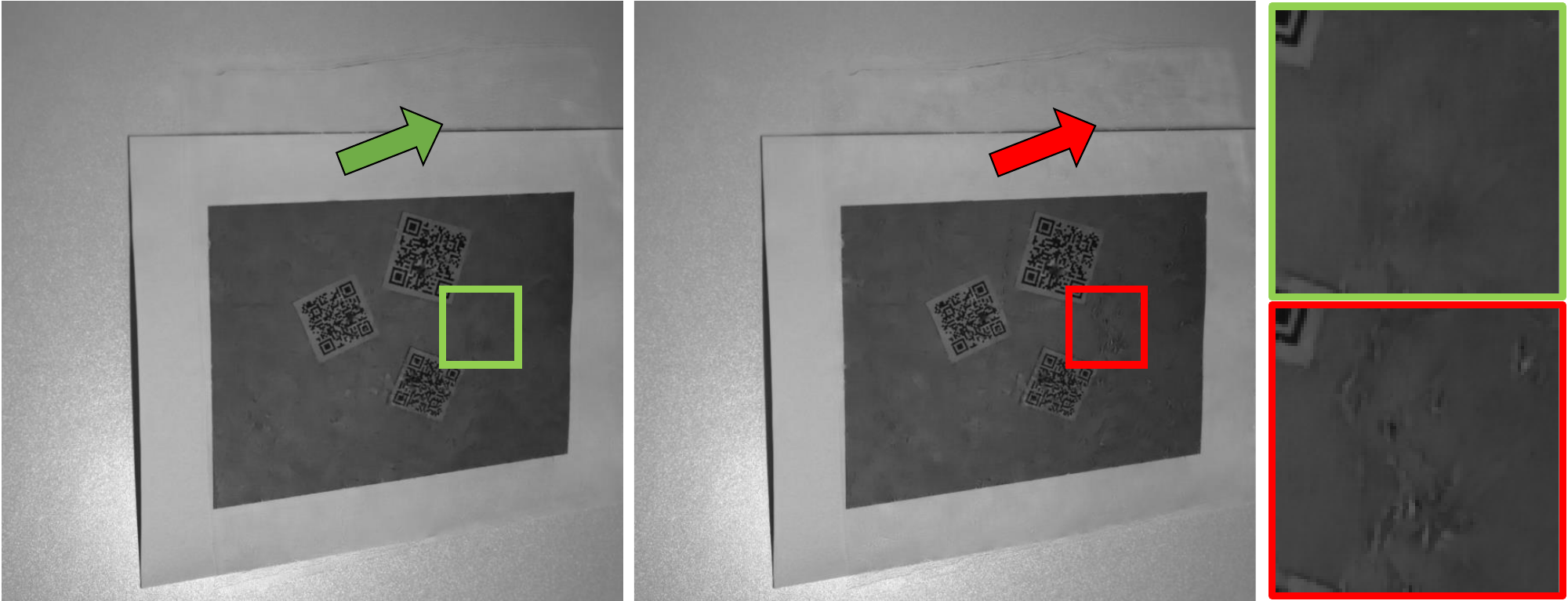} 
	\end{tabular}}
        \vspace{-0.17in}
	\caption{Qualitative evaluation of w/o Polarization Information. Compared with w/o polarization information (\textbf{\textcolor[rgb]{ 1,  0,  0}{Red}}), introducing polarization information avoids the generation of artifacts and reconstructs the realistic details (\textbf{\textcolor[rgb]{ 0,  .69,  .314}{Green}}).}
	\vspace{-0.19in}
	\label{abinfo}
\end{figure}

\noindent\textbf{Effectiveness of the Polarized Information} To prove the effectiveness of polarized information, we eliminate all components that need polarization information, which include four polarization images. Since the Polarization-based Location Model is driven by the $I_{max}$ and $I_{min}$, which need to be calculated with polarization images as input, this structure is also removed.  Other settings remain constant. Without the support of polarized information, the quantitative performance has a considerable decrease, as shown in Fig.~\ref{fig:abpsnr} (w/o Polarized information). Also, the performance of the network is limited by the lack of guidance information, so it cannot remove all degradation. Besides, removing the polarization information will induce artifacts in highlight regions, as shown in Fig.~\ref{abinfo}.

\begin{figure}
	\centering
	\LARGE
	\newcommand\widthface{1.0}
	\resizebox{1.0\linewidth}{!}{
	\begin{tabular}{c}
		\includegraphics[width=\widthface\textwidth]{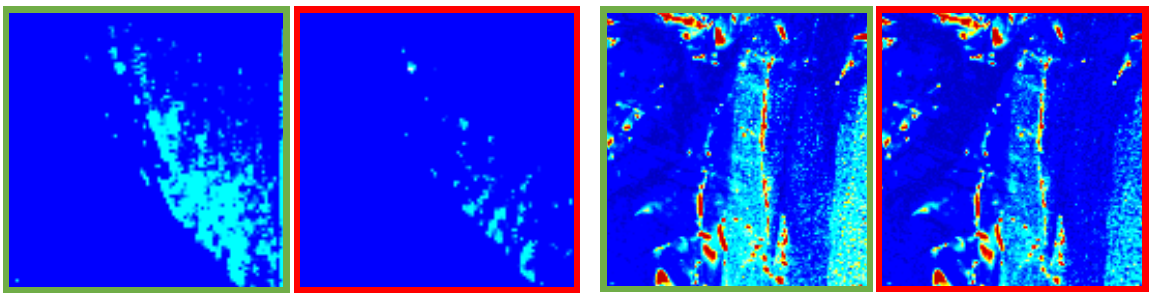}
	\end{tabular}}
        \vspace{-0.21in}
	\caption{Highlight location heatmap. 
 Compared with ``w/o AoP \& DoP'' (In \textbf{\textcolor[rgb]{ 1,  0,  0}{Red} Box}), introducing AoP and DoP (In \textbf{\textcolor[rgb]{ 0,  .69,  .314}{Green} Box}) can help the A-Net to infer more highlight regions.}
	\vspace{-0.14in}
	\label{abpi}
\end{figure}

\noindent\textbf{Effectiveness of AoP and DoP} To prove the effectiveness of AoP and DoP, we eliminate AoP and DoP in the angle estimation network. The qualitative results presented in Fig.~\ref{abpi} demonstrate that AoP and DoP can help locate the highlight more significantly, and also improve overall performance in Fig.~\ref{fig:abpsnr} (w/o AoP \& DoP).

\begin{figure}
	\centering
	\LARGE
        \vspace{-0.05in}
	\newcommand\widthface{1.0}
	\resizebox{1.0\linewidth}{!}{
	\begin{tabular}{c}
	    \includegraphics[width=\widthface\textwidth]{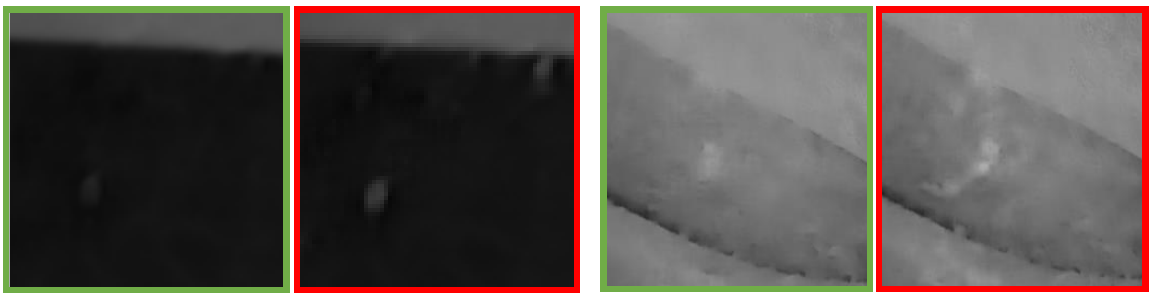}
	\end{tabular}}
        \vspace{-0.2in}
	\caption{Qualitative Evaluation of w/o Prior. Compared with ``w/o Prior'' (In \textbf{\textcolor[rgb]{ 1,  0,  0}{Red} Box}), our solution (In \textbf{\textcolor[rgb]{ 0,  .69,  .314}{Green} Box}) is more effective to remove highlight and infer original information.}
	\vspace{-0.3in}
	\label{ab:prior}
\end{figure}

\noindent\textbf{Effectiveness of Prior} To prove the effectiveness of polarized prior, we eliminate the angle estimation network and the PLM in our network while retaining only the polarized images and the reconstruction network. In this case, the network can only implicitly facilitate the polarized information in the reconstruction network. 
Since there is no polarized prior that indicates highlight regions, 
this can lead to a decrease in the performance of the network to recover highlight regions (as shown in Fig.~\ref{ab:prior}), while causing overall performance to decrease in Fig.~\ref{fig:abpsnr} (w/o Prior).

\vspace{-0.13in}
\section{Conclusion}
\vspace{-0.07in}
\label{sec:conclu}
In this study, we pioneer the investigation of the Film Removal (FR) problem, aiming to eliminate the disturbances caused by wrinkled transparent films and to restore the obscured information. We propose an end-to-end framework to effectively remove all degradations caused by the film with a polarized prior to minimizing highlight. Besides we build a practical polarized dataset containing paired data for this problem. Experiments in the industry have demonstrated the potential application. We believe that the deployment of our algorithms will considerably improve the robustness of downstream industrial recognition systems.

\noindent \textbf{Acknowledgment} This work is supported by the National Natural Science Foundation of China (No. 62206068) and the Natural Science Foundation of Zhejiang Province, China under No. LD24F020002.

{\small
\bibliographystyle{ieee_fullname}
\bibliography{egbib}
}

\end{document}